\documentclass[journal]{IEEEtran}
\usepackage{graphicx}
\usepackage{epsfig}
\usepackage{epstopdf}
\usepackage{float}
\usepackage{amsmath}


\begin{document}
%
\title{Discriminative Multiple Canonical Correlation Analysis for Information Fusion}
%
%
\author{Lei~Gao,~\IEEEmembership{Student Member,~IEEE,}
        Lin~Qi,~
        Enqing~Chen,~\IEEEmembership{Member,~IEEE,}
        and~Ling~Guan,~\IEEEmembership{Fellow,~IEEE}
\thanks{L. Gao is with the Department of Electrical and Computer Engineering, Ryerson University, Toronto, ON M5B 2K3, Canada. (email:iegaolei@gmail.com)}
\thanks{L. Qi and E. Chen are with the School of Information Engineering, Zhengzhou University, Zhengzhou, China. E. Chen is the corresponding author. (email:ielqi@zzu.edu.cn; enqingchen@gmail.com, ieeqchen@zzu.edu.cn)}
\thanks{L. Guan is with the Department of Electrical and Computer Engineering, Ryerson University, Toronto, ON M5B 2K3, Canada (email:lguan@ee.ryerson.ca).}}

\maketitle

\begin{abstract}
 In this paper, we propose the Discriminative Multiple Canonical Correlation Analysis (DMCCA) for multimodal information analysis and fusion. DMCCA is capable of extracting more discriminative characteristics from multimodal information representations. Specifically, it finds the projected directions which simultaneously maximize the within-class correlation and minimize the between-class correlation, leading to better utilization of the multimodal information. In the process, we analytically demonstrate that the optimally projected dimension by DMCCA can be quite accurately predicted, leading to both superior performance and substantial reduction in computational cost. We further verify that Canonical Correlation Analysis (CCA), Multiple Canonical Correlation Analysis (MCCA) and Discriminative Canonical Correlation Analysis (DCCA) are special cases of DMCCA, thus establishing a unified framework for Canonical Correlation Analysis. We implement a prototype of DMCCA to demonstrate its performance in handwritten digit recognition and human emotion recognition. Extensive experiments show that DMCCA outperforms the traditional methods of serial fusion, CCA, MCCA and DCCA.
\end{abstract}

\IEEEpeerreviewmaketitle

\section{Introduction}
%
%
%
%
The effective utilization and integration of multimodal information contents presented in different media sources are becoming an increasingly important research topic in many applications. Since different characteristics of the same data may contain both common and complementary information about the data semantics, the joint use of the multiple characteristics can potentially provide a more discriminatory and complete description of the intrinsic characteristics of the pattern, improving system performance compared with single modality only [1]. It plays important roles in modern multimedia signal processing and is widely applied to human-computer interaction (HCI), human-computer communication (HCC), security/surveillance and many other areas [2-3].\\\indent Information fusion is the study of multiple data modalities such as voice, face, facial expressions, hand and body gesture, ECG and EEG, etc. Among them, voice and face information are two of the most natural, passive, and noninvasive types of traits [4-5]. They can be easily captured by low-cost sensing devices, making them more economically feasible for potential deployment in a wide range of applications. Many methods have been proposed for audio-visual based biometric recognition. Serdar et al. [6] described the use of visual information along with prosodic and language information to detect the presence of disfluencies in a computer-directed speech. Pan et al. [7] presented a novel fused hidden Markov model (fused HMM) for integrating tightly coupled audio and visual features of speech with application to bimodal speech processing. Wang and Guan [8] proposed a hierarchical multiclassifier scheme for audiovisual analysis and achieved the task of recognition of human emotions at the feature level. \\\indent For information fusion, the major difficulties lie in the identification of the discriminatory representations between different modalities, and the design of a fusion strategy that can effectively utilize the complete information presented in different channels. A wide variety of methods have been proposed in the literatures to address the problems. State-of-the-art in multimodal and multimedia related content analysis and recognition are documented in [9-11, 51-54]. Generally, the fusion strategies can be categorized as rule based and machine learning based methods. In the rule based fusion, a predefined rule (e.g., max, min, sum and product) [12] is applied to combine multiple information sources. On the other hand, machine learning based fusion takes the multimodal information into a classification algorithm, which learns the characteristics of the multiple patterns from a set of training samples [12-13].\\\indent It is widely acknowledged that there are three levels of information fusion: feature/data level, score level and decision level [14]. Data/feature level fusion combines the original data or extracted features through certain strategies to form a new vector representing the original information before classification. Fusion at the score level combines the scores generated from multiple classifiers using multiple modalities through a rule based scheme, or in a pattern classification sense in which the scores are taken as new features into a classification algorithm. The decision level fusion usually generates the final results based on the decisions made from individual classifiers or modalities using rule based methods such as AND, OR, and majority voting. Among the three levels fusion, the decision level fusion and score level fusion, delegated by multi-classifier combination, have been researched extensively [15-17]. In recent years, intelligent feature level fusion has drawn significant attention from the research communities of multimedia and biometrics due to its capacity of information preservation and has made impressive progress [18-20]. Among them, serial feature fusion [20] was the early winner. However, after serial feature extraction, how to select low-dimensional discriminative feature vectors for effective recognition remains a challenging open problem. \\\indent Recently, there has been extensive interest in the analysis of correlation based approaches for multimodal information fusion. The objective of correlation analysis is to identify and measure the intrinsic association between different modalities. Hu et al. [47] proposed a large margin multi-metric learning (LM$^3$L) method for face and kinship verification in the wild. Wang et al. [12] proposed the method of kernel cross-modal factor analysis to address the audiovisual emotion recognition problem, achieving promising recognition accuracy on the RML and eNTERFACE datasets with a combination of feature and score level fusions. \\\indent Canonical correlation analysis (CCA) is one of the statistical methods dealing with the mutual relationship between two random vectors, and a valuable means of jointly processing information from bimodal sources. It has been widely applied to data analysis, information fusion and more [21-22]. Sun et al. [23] proposed to use CCA to identify the correlation information of multiple feature streams of an image, and demonstrated the effectiveness of the method in handwriting recognition and face recognition. The CCA method has also been applied to audiovisual based talking-face biometric verification [24], medical imaging analysis [25], and audio-visual synchronization [26]. However, in many practical problems dependencies between two signals cannot be effectively represented by simple linear correlation. If there is nonlinear correlation between the two variables, CCA may not correctly correlate this relationship. \\\indent In order to address the nonlinear problem, two different methods are proposed. Kernel canonical correlation analysis (KCCA) [27], a nonlinear extension of CCA via the kernel trick to overcome this drawback, has been applied to the fusion of global and local features for target recognition [28], the fusion of ear and profile face for multimodal biometric recognition [29], the fusion of text and image for spectral clustering [30], and the fusion of labelled graph vector and the semantic expression vector for facial expression recognition [31]. On the other hand, Deep Canonical Correlation Analysis (Deep CCA) was introduced in [48]. It attempts to learn complex nonlinear transformations of two views of data by neural networks such that the resulting representations were highly linearly correlated. \\\indent However, by CCA, KCCA or Deep CCA, only the correlation between the pairwise samples is better revealed. This correlation can neither well represent the similarity between the samples in the same class, nor can it evaluate the dissimilarity between the samples in different classes. To tackle the problem, a new supervised learning method, namely discriminative CCA (DCCA) is proposed [32-34]. It can simultaneously maximize the within-class correlation and minimize the between-class correlation, thus potentially more suitable for recognition tasks than CCA, KCCA or Deep CCA.  \\\indent Nevertheless, CCA, KCCA, DCCA and Deep CCA only deal with the mutual relationship between two random vectors, limiting the application of the technique if there are multiple random vectors. Multi-set canonical correlation analysis (MCCA) is a natural extension of two-set canonical correlation analysis. It is generalized from CCA to deal with multiple sets of features. The idea is to optimize characteristics of the dispersion matrix of the transformed variables to obtain high correlations between all new variables simultaneously. It has been applied for inclusion in geographical information systems (GIS) [35], joint blind source separation [36] and blind single-input $ \& $ multiple-output (SIMO) channels equalization [37]. However, it does not explore the discriminatory representation and is not capable of providing satisfactory recognition performance.\\\indent In this paper, we introduce a discriminative multiple canonical correlation analysis (DMCCA) approach for multimodal analysis and fusion. We will mathematically verify the following characteristics of DMCCA: a) The optimally projected dimensions can be accurately predicted, leading to both superior performance and substantial reduction in computational cost; b) CCA, DCCA and MCCA are special cases of DMCCA, thus establishing a unified framework for canonical correlation analysis with the purpose of information fusion in the transformed domain. The effectiveness of DMCCA is demonstrated and compares with serial fusion and methods based on similar principles such as CCA, DCCA and MCCA with applications to handwritten digit recognition and human emotion recognition problems.\\\indent The remainder of this paper is organized as follows: the analysis and derivation of DMCCA are introduced in Section II. In Section III, feature extraction and implementation of DMCCA for multimodal fusion with application to handwritten digit recognition and human emotion recognition are presented. The experimental results and analysis are given in Section IV. Conclusions are drawn in Section V.
\section{The DMCCA Method}
One of the major challenges in information fusion is to identify the discriminatory representation amongst different modalities. In this section, we introduce discriminative multiple canonical correlation analysis (DMCCA) to address this problem. Although Generalized multi-view analysis (GMA) [55] and Multi-view Discriminant Analysis (MDA) [56] are also proposed to solve the multi-view problem, there exist obvious differences among them. To be specific, the differences between DMCCA and GMA are: 1) In GMA, the discriminability is obtained within each feature, while in DMCCA it is achieved by using all features; 2) In GMA, the cross-view correlation is obtained only from observations corresponding to the same underlying sample, while in DMCCA it is obtained from all observations from different feature sets; 3) GMA needs a significant number of processing parameters, especially when the feature vector is long, while DMCCA works with a much smaller set of parameters. \\\indent Simultaneously, different from MDA, the purpose of DMCCA is to simultaneously maximize the within-class correlation and minimize the between-class correlation, helping reveal the intrinsic structure and complementary representations from different sources/modalities in order to improve the recognition accuracy. \\\indent The advantages of DMCCA for multimodal information fusion rest on the following facts: 1) DMCCA involves modalities/features having a mixture of correlated (modality-common information) components and achieving the maximum of the correlation [57]. Therefore, DMCCA possesses the maximal commonality of multiple modalities/features; 2) the within-class and the between-class correlations of all modalities/features are considered jointly to extract more discriminative information, leading to a more discriminant common space and better generalization ability for classification from multiple modalities/features. In the following, we first briefly present the fundamentals of CCA, DCCA and MCCA, and then formulate DMCCA.


\subsection{Review of CCA, DCCA and MCCA}

\subsubsection{CCA}

The aim of CCA is to find basis vectors for two sets of variables such that the correlations between the projections of the variables onto these basis vectors are mutually maximized. Simultaneously, it needs to satisfy the canonical property that the first projection is uncorrelated with the second projection, etc. To do so, all useful information, be it common to the two sets or specific to one of them, is maximumly preserved through the projections.\\\indent Let $ x \in {R^{\;m}},y \in {R^{\;p}} $ be two sets as the entries, and \emph{$ m $}, \emph{$ p $} being the dimensions of the samples in \emph{$ x $} and \emph{$ y $}, respectively. The CCA finds a pair of directions $ \omega_1 $ and $ \omega_2 $ to maximize the correlation between the projections of the two canonical vectors: $ X = \omega_1^Tx  $, $ Y = \omega_2^Ty  $, which can be written as follows:
\begin{equation} \mathop {\arg \max }\limits_{{\omega _1},{\omega _2}}  {\omega _1}^T{R_{xy}}{\omega _2}\end{equation} \\ where
${R_{xy}} = x{y^T}$ is the cross-correlation matrix of the vectors $ x $ and $ y $. \\
Simultaneously, $ x $ and $ y $ should satisfy the following constrained condition to guarantee the first projection is uncorrelated with the second projection(canonical property):
\begin{equation} {\omega _1}^T{R_{xx}}{\omega _1} = {\omega _2}^T{R_{yy}}{\omega _2} = 1 \end{equation} \\ By solving the above problem using the algorithm of Lagrange multipliers, we obtain the following relationship [26]:
\begin{equation} \left[ \begin{array}{l}
 0{\rm{        }} \\
 {R_{yx}}{\rm{     }} \\
 \end{array} \right.\left. \begin{array}{l}
 {{\rm{R}}_{xy}} \\
 {\rm{0}} \\
 \end{array} \right]\omega {\rm{ = }}\mu \left[ \begin{array}{l}
 {{\rm{R}}_{xx}}{\rm{        }} \\
 {\rm{   0  }} \\
 \end{array} \right.\left. \begin{array}{l}
 0 \\
 {R_{yy}} \\
 \end{array} \right]\omega
 \end{equation} \\ where $ \mu $ is the canonical correlation value and $\omega  = {[{\omega _1^T},{\omega _2^T}]^T}$ is the projected vector. Then equation (3) can be solved as the generalized eigenvalue(GEV) problem.
 \subsubsection{DCCA}
The purpose of DCCA is to maximize the similarities of any pairs of sets of within-class while minimizing the similarities of pairwise sets of between-class, as mathematically expressed in [33]:
\begin{equation}
T = {\max _{\arg T}}tr({T^{'}}{S_w}T)/tr({T^{'}}{S_b}T)
\end{equation}
where \emph{T} is the discriminant function and $S_w, S_b$ relate to the within-class and between-class matrices respectively.\\The solution to equation (4) is obtained by solving the following generalized eigenvalue problem:
\begin{equation}
{S_b}T = \lambda {S_w}T
\end{equation}
For detailed information, please refer to [33].
\subsubsection{MCCA}
  MCCA can be viewed as a natural extension of the two-set canonical correlation analysis [37]. Given $ M $ sets of real random variables ${x_1},{\rm{ }}{x_2}, \cdots {x_M}$ with the dimensions of ${m_1},{\rm{ }}{m_2}, \cdots {m_M}$. The objective of MCCA is to find $\omega  = {[{\omega _1}^T,{\omega _2}^T \cdots {\omega _M}^T]^T}$ which satisfies similar requirement as CCA and described as: \\
 \begin{equation} \mathop {\arg \max }\limits_{{\omega _1},{\omega _2} \cdots {\omega _M}}  \frac{1}{{M(M - 1)}}\sum\limits_{\scriptstyle k,l = 1 \hfill \atop
  \scriptstyle {\rm{ }}k \ne l \hfill}^M {{\omega _k}^T{C_{{x_k}{x_l}}}{\omega _l}} {\rm{  (}}k \ne l{\rm{)}} \end{equation}\\
  subject to \begin{equation} \sum\limits_{k = 1}^M {{\omega _k}^T{C_{{x_k}{x_k}}}{\omega _k}}  = M \end{equation}\\
  where ${C_{{x_k}{x_l}}} = {x_k}{x_l}^T$. Equation (6) can be solved by the Lagrange multipliers as: \\
  \begin{equation} \frac{1}{{M - 1}}(C - D)\omega  = \beta D\omega \end{equation} \\ where \\
\begin{equation}\ C = \left[ {\begin{array}{*{20}{c}}
   {{x_1}{x_1}^T} &  \ldots  & {{x_1}{x_M}^T}  \\
    \vdots  &  \ddots  &  \vdots   \\
   {{x_M}{x_1}^T} &  \cdots  & {{x_M}{x_M}^T}  \\
\end{array}} \right]
 \end{equation} \\
 \begin{equation}\ D = \left[ {\begin{array}{*{20}{c}}
   {{x_1}{x_1}^T} &  \ldots  & 0  \\
    \vdots  &  \ddots  &  \vdots   \\
   0 &  \cdots  & {{x_M}{x_M}^T}  \\
\end{array}} \right]
\end{equation} \\with $ \beta $ being the generalized canonical correlation. The MCCA solution is obtained as the GEV problem.
\subsection{Concept of the DMCCA}
Let \emph{$ P $} sets of zero-mean random variables be ${x_1} \in {R^{{m_1}}},{x_2} \in {R^{{m_2}}}, \cdots {x_P} \in {R^{{m_P}}}$ for \emph{c} classes and $Q = {m_1} + {m_2} +  \cdots {m_P}$. Concretely, DMCCA aims to seek the projection vectors $\omega  = {[{\omega _1}^T,{\omega _2}^T \cdots {\omega _P}^T]^T} ({\omega _1} \in {R^{{m_1} \times Q}},{\omega _2} \in {R^{{m_2} \times Q}}, \cdots {\omega _P} \in {R^{{m_P} \times Q}})$ for information fusion so that the within-class correlation is maximized and the between-class correlation is minimized. Based on the definition of CCA and MCCA, DMCCA is formulated as the following optimization problem:
\begin{equation} \mathop {\arg \max \rho }\limits_{{\omega _1},{\omega _2} \cdots {\omega _P}}  = \frac{1}{{P(P - 1)}}\sum\limits_{\scriptstyle k,m = 1\hfill\atop \scriptstyle{\rm{ }}k \ne m\hfill}^P {{\omega _k}^T\mathop {{C_{{x_k}{x_m}}}}\limits^ \sim  {\omega _m}} {\rm{  (}}k \ne m{\rm{)}} \end{equation}
Subject to
\begin{equation} \sum\limits_{k = 1}^P {{\omega _k}^T{C_{{x_k}{x_k}}}{\omega _k}}  = P \end{equation}
where $\mathop {{C_{{x_k}{x_m}}}}\limits^ \sim   = {C_{{w_{{x_k}{x_m}}}}} - \delta {C_{{b_{{x_k}{x_m}}}}}(\delta  > 0),{\rm{ }}{C_{{x_k}{x_k}}} = {x_k}{x_k}^T$. ${C_{{w_{{x_k}{x_m}}}}}$ and ${C_{{b_{{x_k}{x_m}}}}}$ denote the within-class and between-class correlation matrixes of \emph{$ P $} sets, respectively. \\\indent
Based on the mathematical analysis in Appendix \emph{I}, $C_{{w_{{x_k}{x_m}}}}$ and $C_{{b_{{x_k}{x_m}}}}$ can be explicitly expressed in equations (13), (14) and (15).
\begin{equation}
{C_{{w_{{x_k}{x_m}}}}} = {x_k}A{x_m}^T
\end{equation}
\begin{equation}
{C_{{b_{{x_k}{x_m}}}}} = - {x_k}A{x_m}^T
\end{equation}
\begin{equation} \ A = \left[ {\left( {\begin{array}{*{20}{c}}{{H_{{n_{i1}} \times {n_{i1}}}}}& \ldots &0\\
 \vdots &{{H_{{n_{il}} \times {n_{il}}}}}& \vdots \\
0& \ldots &{{H_{_{{n_{ic}} \times {n_{ic}}}}}}
\end{array}} \right)} \right]  \end{equation}
where ${n_{il}}$ is the number of samples in the \emph{l}th class of ${x_i}$ set and ${H_{{n_{i{l}}} \times {n_{il}}}}$ is in the form of ${n_{il}} \times {n_{il}}$ and all the elements in ${H_{{n_{i{l}}} \times {n_{il}}}}$ are unit values.\\
Substituting equations (13) and (14) into (11) yields:
\begin{equation} \begin{array}{l}
\mathop {\arg \max \rho }\limits_{{\omega _1},{\omega _2} \cdots {\omega _P}}  = \frac{1}{{P(P - 1)}}\sum\limits_{k,m = 1}^P {{\omega _k}^T\mathop {{C_{{x_k}{x_m}}}}\limits^ \sim  {\omega _m}} {\rm{  }}\\
{\rm{ = }}\frac{{1 + \delta }}{{P(P - 1)}}\sum\limits_{k,m = 1}^P {{\omega _k}^T{x_k}A{x_m}^T{\omega _m}}
\end{array} \end{equation}
Subject to
\begin{equation} \sum\limits_{k = 1}^P {{\omega _k}^T{C_{{x_k}{x_k}}}{\omega _k}}  = P \end{equation}
Using Lagrangian multiplier criterion, it obtains equation (18) as follows:
\begin{scriptsize}
\begin{equation}
\begin{array}{l}
\frac{{1 + \delta }}{{P - 1}}\left[ {\left( {\begin{array}{*{20}{c}}
0&{{x_1}A{x_2}^T}&{{x_1}A{x_3}^T \ldots }&{{x_1}A{x_P}^T}\\
{{x_2}A{x_1}^T}&0&{{x_2}A{x_3}^T \ldots }&{{x_2}A{x_P}^T}\\
\begin{array}{l}
{x_3}A{x_1}^T\\
{\rm{    }}
\end{array}&\begin{array}{l}
{x_3}A{x_2}^T\\
{\rm{     }} \vdots
\end{array}&\begin{array}{l}
{\rm{    }}0{\rm{  }} \ldots \\
{\rm{    }}
\end{array}&\begin{array}{l}
{x_3}A{x_P}^T\\
{\rm{   }}
\end{array}\\
{{x_P}A{x_1}^T}&{{x_P}A{x_2}^T}&{{x_P}A{x_3}^T \ldots }&0
\end{array}} \right)} \right]\omega \\
 = \rho \left( {\begin{array}{*{20}{c}}
{{x_1}{x_1}^T}&0&{0 \ldots }&0\\
0&{{x_2}{x_2}^T}&{0 \ldots }&0\\
\begin{array}{l}
{\rm{ }}0\\
{\rm{   }}
\end{array}&\begin{array}{l}
{\rm{  }}0\\
{\rm{     }} \vdots
\end{array}&\begin{array}{l}
{\rm{    }}{x_3}{x_3}^T \ldots \\
{\rm{    }}
\end{array}&\begin{array}{l}
{\rm{ }}0\\
{\rm{   }}
\end{array}\\
0&0&{0 \ldots }&{{x_P}{x_P}^T}
\end{array}} \right)\omega
\end{array}
\end{equation}
\end{scriptsize}

It is further rewritten as:
\begin{equation} \frac{{1+ \delta}}{{P - 1}}(C - D)\omega  = \rho D\omega \end{equation}
where
\begin{small}
\begin{equation}
\;C = \left[ {\left( {\begin{array}{*{20}{c}}
{{x_1}{x_1}^T}&{{x_1}A{x_2}^T}&{{x_1}A{x_3}^T \ldots }&{{x_1}A{x_P}^T}\\
{{x_2}A{x_1}^T}&{{x_2}{x_1}^T}&{{x_2}A{x_3}^T \ldots }&{{x_2}A{x_P}^T}\\
\begin{array}{l}
{x_3}A{x_1}^T\\
{\rm{    }}
\end{array}&\begin{array}{l}
{x_3}A{x_2}^T\\
{\rm{     }} \vdots
\end{array}&\begin{array}{l}
{\rm{    }}{x_3}{x_3}^T{\rm{  }} \ldots \\
{\rm{    }}
\end{array}&\begin{array}{l}
{x_3}A{x_P}^T\\
{\rm{   }}
\end{array}\\
{{x_P}A{x_1}^T}&{{x_P}A{x_2}^T}&{{x_P}A{x_3}^T \ldots }&{{x_P}{x_P}^T}
\end{array}} \right)} \right]
\end{equation}
\end{small}
\begin{small}
\begin{equation}
D = \left[ {\left( {\begin{array}{*{20}{c}}
{{x_1}{x_1}^T}&0&{0 \ldots }&0\\
0&{{x_2}{x_2}^T}&{0 \ldots }&0\\
\begin{array}{l}
{\rm{ }}0\\
{\rm{   }}
\end{array}&\begin{array}{l}
{\rm{  }}0\\
{\rm{     }} \vdots
\end{array}&\begin{array}{l}
{\rm{    }}{x_3}{x_3}^T \ldots \\
{\rm{    }}
\end{array}&\begin{array}{l}
{\rm{ }}0\\
{\rm{   }}
\end{array}\\
0&0&{0 \ldots }&{{x_P}{x_P}^T}
\end{array}} \right)} \right]
\end{equation}
\end{small}
\begin{small}
\begin{equation}
\:C - D = \left( {\begin{array}{*{20}{c}}
   0 & {{x_1}A{x_2}^T} & {{x_1}A{x_3}^T \ldots } & {{x_1}A{x_P}^T}  \\
   {{x_2}A{x_1}^T} & 0 & {{x_2}A{x_3}^T \ldots } & {{x_2}A{x_P}^T}  \\
   {\begin{array}{*{20}{c}}
   {{x_3}A{x_1}^T}  \\
   {}  \\
\end{array}} & {\begin{array}{*{20}{c}}
   {{x_3}A{x_2}^T}  \\
    \vdots   \\
\end{array}} & {\begin{array}{*{20}{c}}
   {0 \ldots }  \\
   {}  \\
\end{array}} & {\begin{array}{*{20}{c}}
   {{x_3}A{x_P}^T}  \\
   {}  \\
\end{array}}  \\
   {{x_P}A{x_1}^T} & {{x_P}A{x_2}^T} & {{x_P}A{x_3}^T \ldots } & 0  \\
\end{array}} \right)
\end{equation}
\end{small}
\begin{small}\begin{equation} \omega  = [{\omega ^T}_1,{\omega ^T}_2, \cdots {\omega ^T}_P]^T \end{equation}\end{small}
Since $({{1 + \delta } \mathord{\left/
 {\vphantom {{1 + \delta } {P - 1}}} \right.
 \kern-\nulldelimiterspace} {P - 1}})$ is a constant, it will not introduce any influence to the projection matrix $\omega$. During the following mathematical analysis, we will ignore this term. With mathematical transform, equation (19) is further written in the form of:
\begin{small}
\begin{equation}
\begin{array}{l}
{x_1}A{x_2}^T{\omega _2} + {x_1}A{x_3}^T{\omega _3} +  \cdots  + {x_1}A{x_P}^T{\omega _P} = {\rho}{x_1}{x_1}^T{\omega _1}\\
{x_2}A{x_1}^T{\omega _1} + {x_2}A{x_3}^T{\omega _3} +  \cdots  + {x_2}A{x_P}^T{\omega _P} = {\rho}{x_2}{x_2}^T{\omega _2}\\
                \centerline \vdots \\
{x_P}A{x_1}^T{\omega _1} + {x_P}A{x_2}^T{\omega _2} +  \cdots  + {x_P}A{x_{P - 1}}^T{\omega _{P - 1}} = {\rho}{x_P}{x_P}^T{\omega _P}
\end{array}
\end{equation}
\end{small}\\\indent
Based on the definition of $\mathop {{C_{{x_k}{x_m}}}}\limits^ \sim  $ and equation (19), the value of ${\rho}$ plays a critical role in evaluating the relationship between within-class and between-class correlation matrixes. When the value of ${\rho}$ is greater than zero, the corresponding projected vector ${\omega}$ contributes positively to the discriminative power in classification while the projected vector ${\omega}$ corresponding to the non-positively values of ${\rho}$ would result in reducing the discriminative power in classification. Clearly, the solution obtained is the eigenvector associated to the largest eigenvalue in equation (19).\\\indent
Commonly, it is known that the time taken greatly depends on the computational process of projective vectors, using algebraic method to extract discriminative features. When the rank of eigen-matrix is very high, the computation of eigenvalues and eigenvectors will be time-consuming. To address this problem effectively, an important property of DMCCA is proved here. That is the number of projected dimension \emph{d} corresponding to the optimal recognition accuracy is smaller than or equals to the number of classes, \emph{c}, or mathematically:
\begin{equation} \ d \le c \end{equation}\\\indent
Now we will show that \emph{d} does satisfy inequality (25).
From equation (15), the rank of matrix \emph{A} satisfies
\begin{equation} \ rank(A) \le c \end{equation}\\\indent
Then, equation (26) leads to:
\begin{equation} \ rank({x_i}A{x_j}^T) \le \min ({r_i},{r_A},{r_j}) \end{equation}
where ${r_i},{\rm{  }}{r_A},{\rm{  }}{r_j}$ are the ranks of matrixes ${x_i},A,{x_j}$ (
$ i,j \in [1,2,3,...,P] $), respectively.\\\indent
Due to the fact that $rank(A) \le c$, equation (27) satisfies
\begin{equation} \ rank({x_i}A{x_j}^T) \le \min ({r_i},{c},{r_j}) \end{equation}
when $c$ is less than ${r_i}$ and ${r_j}$, equation (28) is written as
\begin{equation} \ rank({x_i}A{x_j}^T) \le {c} \end{equation}
otherwise, equation (28) satisfies
\begin{equation} \ rank({x_i}A{x_j}^T) \le min({r_i,r_j}) <\ {c} \end{equation}\\\indent
Then equation (24) can be written as follows:
\begin{small}
\begin{equation}
\left\{ {\begin{array}{*{20}{c}}
   {{x_1}A({x_2}^T{\omega _2} + {x_3}^T{\omega _3} +  \cdots  + {x_P}^T{\omega _P}) = {\rho}{x_1}{x_1}^T{\omega _1}}  \\
   {{x_2}A({x_1}^T{\omega _1} + {x_3}^T{\omega _3} +  \cdots  + {x_P}^T{\omega _P}) = {\rho}{x_2}{x_2}^T{\omega _2}}  \\
    \vdots   \\
   {{x_P}A({x_1}^T{\omega _1} + {x_2}^T{\omega _2} +  \cdots  + {x_{P - 1}}^T{\omega _{P - 1}}) = {\rho}{x_P}{x_P}^T{\omega _P}}  \\
\end{array}} \right.
\end{equation}
\end{small}
It is further written as:
\begin{equation}
\begin{array}{l}
 \left[ {\begin{array}{*{20}{c}}
   {{x_1}} & 0 & {0,...} & 0  \\
   0 & {{x_2}} & {0,...} & 0  \\
   {} &  \vdots  & {} & {}  \\
   0 & 0 & {0,...} & {{x_P}}  \\
\end{array}} \right]\underbrace {\left[ {\begin{array}{*{20}{c}}
   A & 0 & {0,...} & 0  \\
   0 & A & {0,...} & 0  \\
   {} &  \vdots  & {} & {}  \\
   0 & 0 & {0,...} & A  \\
\end{array}} \right]}_P\left[ {\begin{array}{*{20}{c}}
   {{B_1}}  \\
   {{B_2}}  \\
    \vdots   \\
   {{B_P}}  \\
\end{array}} \right] \\
  = \rho \left[ {\begin{array}{*{20}{c}}
   {{x_1}{x_1}^T} & 0 & {0,...} & 0  \\
   0 & {{x_2}{x_2}^T} & {0,...} & 0  \\
   {} &  \vdots  & {} & {}  \\
   0 & 0 & {0,...} & {{x_P}{x_P}^T}  \\
\end{array}} \right]\left[ {\begin{array}{*{20}{c}}
   {{\omega _1}}  \\
   {{\omega _2}}  \\
    \vdots   \\
   {{\omega _P}}  \\
\end{array}} \right] \\
 \end{array}
\end{equation}
where
\begin{equation}
\left\{ {\begin{array}{*{20}{c}}
   {{B_1} = {x_2}^T{\omega _2} + {x_3}^T{\omega _3} +  \cdots  + {x_P}^T{\omega _P}}  \\
   {{B_2} = {x_1}^T{\omega _1} + {x_3}^T{\omega _3} +  \cdots  + {x_P}^T{\omega _P}}  \\
    \vdots   \\
   {{B_P} = {x_1}^T{\omega _1} + {x_2}^T{\omega _2} +  \cdots  + {x_{P - 1}}^T{\omega _{P - 1}}}  \\
\end{array}} \right.
\end{equation}\\\indent
Equation (31) is further expressed as:
\begin{equation}
\left\{ {\begin{array}{*{20}{c}}
   {{{({\zeta _1})}^{ - 1}}{x_1}A({x_2}^T{\omega _2} + {x_3}^T{\omega _3} +  \cdots  + {x_P}^T{\omega _P}) = {\omega _1}}  \\
   {{{({\zeta _2})}^{ - 1}}{x_2}A({x_1}^T{\omega _1} + {x_3}^T{\omega _3} +  \cdots  + {x_P}^T{\omega _P}) = {\omega _2}}  \\
    \vdots   \\
   {{{({\zeta _P})}^{ - 1}}{x_P}A({x_1}^T{\omega _1} + {x_2}^T{\omega _2} +  \cdots  + {x_{P - 1}}^T{\omega _{P - 1}}) = {\omega _P}}  \\
\end{array}} \right.
\end{equation}
where
\begin{equation}
\left\{ {\begin{array}{*{20}{c}}
   {{\zeta _1} = {\rho}{x_1}{x_1}^T}  \\
   {{\zeta _2} = {\rho}{x_2}{x_2}^T}  \\
    \vdots   \\
   {{\zeta _P} = {\rho}{x_P}{x_P}^T}  \\
\end{array}} \right.(when\;{\rm{ }}{x_i}{x_i}^T{\rm{ \;is\; non-singular}})
\end{equation}
or\\
\begin{equation}
\left\{ {\begin{array}{*{20}{c}}
   {{\zeta _1} = {\rho}{x_1}{x_1}^T + {\sigma _1}{I_1}}  \\
   {{\zeta _2} = {\rho}{x_2}{x_2}^T + {\sigma _2}{I_2}}  \\
    \vdots   \\
   {{\rm{  }}{\zeta _P} = {\rho}{x_P}{x_P}^T + {\sigma _P}{I_P}}  \\
\end{array}} \right.(when\;{x_i}{x_i}^T\;is\;{\rm{singular}})
\end{equation}
where ${I_i} \in {R^{{m_i} \times {m_i}}}$ and ${\sigma _1},{\sigma _2},...{\sigma _P}$ are constants.\\\indent
Since $rank(A) \le c$ and ${\omega _i} \in {R^{{m_i} \times Q}}(i = 1,2,...P)$, based on equation (34), the rank of ${\omega _i}$ satisfies
\begin{equation}
rank({\omega _i}) \le c{\rm{    }}      (i = 1,2,...,P)
\end{equation}\\\indent
Then the fused feature of ${Y_i}$ can be written as follows:
\begin{equation}
{Y_i} = {\omega _i}^T{x_i}{\rm{    }}(i = 1,2,...P)
\end{equation}
leading to the expression of the fused features as:
\begin{equation}
\left\{ {\begin{array}{*{20}{c}}
   {{Y_1} = {\omega _1}^T{x_1}}  \\
   {{Y_2} = {\omega _2}^T{x_2}}  \\
    \vdots   \\
   {{Y_P} = {\omega _P}^T{x_P}}  \\
\end{array}} \right.
\end{equation}\\\indent
Let ${(\omega )_d}$ be the projected matrix with DMCCA achieving optimal performance and ${(\omega )_d} \in {R^{Q \times d}}$. ${(\omega )_d}$ is written as follows£º
\begin{equation}
{(\omega )_d}{\rm{ = }}\left[ {\begin{array}{*{20}{c}}
   {{\omega _{11}},{\omega _{12}},...,{\omega _{1d}}}  \\
   {{\omega _{21}},{\omega _{22}},...,{\omega _{2d}}}  \\
    \vdots   \\
   {{\omega _{P1}},{\omega _{P2}},...,{\omega _{Pd}}}  \\
\end{array}} \right]{\rm{ = }}\left[ {\begin{array}{*{20}{c}}
   {{{({\omega _1})}_d}}  \\
   {{{({\omega _2})}_d}}  \\
    \vdots   \\
   {{{({\omega _P})}_d}}  \\
\end{array}} \right]
\end{equation}\\\indent
Then ${\omega}$, ${(\omega )_d}$ and ${({\omega _i})_d}(i = 1,2,...P)$ satisfy the relationship:
\begin{equation}
\begin{array}{l}
 \omega  = \left[ {\begin{array}{*{20}{c}}
   {{\omega _1}}  \\
   {{\omega _2}}  \\
    \vdots   \\
   {{\omega _P}}  \\
\end{array}} \right]{\rm{ = }}\left[ {\begin{array}{*{20}{c}}
   {{\omega _{11}}} & {{\omega _{12}}} & {...} & {{\omega _{1d}}} & {{\omega _{1(d + 1)}}} & {...} & {{\omega _{1Q}}}  \\
   {{\omega _{21}}} & {{\omega _{22}}} & {...} & {{\omega _{2d}}} & {{\omega _{2(d + 1)}}} & {...} & {{\omega _{2Q}}}  \\
   {} & {} & {} &  \vdots  & {} & {...} & {}  \\
   {{\omega _{P1}}} & {{\omega _{P2}}} & {...} & {{\omega _{Pd}}} & {{\omega _{P(d + 1)}}} & {...} & {{\omega _{PQ}}}  \\
\end{array}} \right] \\
  = \left[ {\begin{array}{*{20}{c}}
   {{{({\omega _1})}_d},{\omega _{1(d + 1)}},...,{\omega _{1Q}}}  \\
   {{{({\omega _2})}_d},{\omega _{2(d + 1)}},...,{\omega _{2Q}}}  \\
    \vdots   \\
   {{{({\omega _P})}_d},{\omega _{P(d + 1)}},...,{\omega _{PQ}}}  \\
\end{array}} \right] \\
  = {\rm{[(}}\omega {)_d},\underbrace {0,...,0}_{Q - d}{\rm{] + [}}\omega {\rm{ - [(}}\omega {)_d},\underbrace {0,...,0}_{Q - d}{\rm{]]   (}}0 \in {R^Q}{\rm{)}} \\
 \end{array}
\end{equation}
Inserting (41) into (39) yields:
\begin{small}
\begin{equation}
\left\{ {\begin{array}{*{20}{c}}
   {{Y_1} = {\omega _1}^T{x_1} = {{{\rm{[(}}{\omega _1}{)_d},\underbrace {{0_1},...,{0_1}}_{Q - d}{\rm{]}}}^T}{x_1} + {{{\rm{[}}{\omega _1}{\rm{ - [(}}{\omega _1}{)_d},\underbrace {{0_1},...,{0_1}}_{Q - d}{\rm{]]}}}^T}{x_1}}  \\
   {{Y_2} = {\omega _2}^T{x_2}{\rm{ = [(}}{\omega _2}{)_d},\underbrace {{0_2},...,{0_2}}_{Q - d}{{\rm{]}}^T}{x_2} + {{{\rm{[}}{\omega _2}{\rm{ - [(}}{\omega _2}{)_d},\underbrace {{0_2},...,{0_2}}_{Q - d}{\rm{]]}}}^T}{x_2}}  \\
    \vdots   \\
   {{Y_P} = {\omega _P}^T{x_P}{\rm{ = [(}}{\omega _P}{)_d},\underbrace {{0_P},...,{0_P}}_{Q - d}{{\rm{]}}^T}{x_P} + {{{\rm{[}}{\omega _P}{\rm{ - [(}}{\omega _P}{)_d},\underbrace {{0_P},...,{0_P}}_{Q - d}{\rm{]]}}}^T}{x_P}}  \\
\end{array}} \right.
\end{equation}
\end{small}
where ${0_i}(i = 1,2,...,P)$ is in the form ${R^{{m_i}}}$.\\\indent
Based on equation (39), $d$ should satisfy inequality (43)
\begin{equation}
d \le \min ({m_1},{m_2},...,{m_P})
\end{equation}
Simultaneously, since $rank({\omega _i}) \le c$ and ${Y_i}$ possesses the same number of rows, $d$ satisfies the relation (44)                                                                                                      \begin{equation}
d \le min[rank({\omega _1}),rank({\omega _2}),...,rank({\omega _P})] \le c
\end{equation}
considering (43) and (44) together, $d$ should satisfy (45)
\begin{equation}
\left\{ \begin{array}{l}
 d \le \min ({m_1},{m_2},...,{m_P}) \\
 d \le c \\
 \end{array} \right.
\end{equation}\\\indent
Now, we analyze the following two cases.\\
\textbf{1. when $c$ satisfies} (46)
\begin{equation}
c \le \min ({m_1},{m_2},...,{m_P})
 \end{equation}
combining (43) and (44) leads to
\begin{equation}
d \le c
 \end{equation}
\textbf{2. when $c$ satisfies} (48)
\begin{equation}
c > \min ({m_1},{m_2},...,{m_P})
\end{equation}
combining (43) and (44) leads to
\begin{equation}
d \le \min ({m_1},{m_2},...,{m_P}) \le c
\end{equation}
In summary,
\begin{equation}
d \le c
\end{equation}\\\indent
In order to achieve the optimal recognition accuracy, we select the $c$ projected vectors from the eigenvectors associated with the $c$ different largest eigenvalues in equation (19).\\\indent
Since ${{\rm{[}}{\omega _i}{\rm{ - [(}}{\omega _i}{)_d},\underbrace {{0_i},...,{0_i}}_{Q - d}{\rm{]]}}^T}{x_i}$
 and ${[\underbrace {{0_i},...,{0_i}}_{Q - d}{\rm{]}}^T}{x_i}$
 have no contribution to the optimally fused result of $Y_i$, the optimal performance reached by DMCCA when, $d$, the projected dimension is less than or equals to $c$, as expressed in (51)
\begin{small}
\begin{equation}
\left\{ {\begin{array}{*{20}{c}}
   {{Y_{1,optimal}}{\rm{ = [(}}{\omega _1}{)_d},\underbrace {{0_1},...,{0_1}}_{Q - d}{{\rm{]}}^T}{x_1} = {{{\rm{(}}{\omega _1})}_d}^T{x_1} \in {R^d}(d <  = c)}  \\
   {{Y_{2,optimal}} = {{{\rm{[(}}{\omega _2}{)_d},\underbrace {{0_2},...,{0_2}}_{Q - d}{\rm{]}}}^T}{x_2} = {{{\rm{(}}{\omega _2})}_d}^T{x_2} \in {R^d}(d <  = c)}  \\
    \vdots   \\
   {{Y_{P,optimal}} = {{{\rm{[(}}{\omega _P}{)_d},\underbrace {{0_P},...,{0_P}}_{Q - d}{\rm{]}}}^T}{x_P} = {{{\rm{(}}{\omega _P})}_d}^T{x_P} \in {R^d}(d <  = c)}  \\
\end{array}} \right.
\end{equation}
\end{small}
Thus, expressions in (51) lead to the proof of (25).\\\indent Specifically, if the feature space dimension equals \emph{Q}, the computational complexity of DMCCA is on the order of \emph{O(Q*c)}, instead of \emph{O(Q*Q)}, as other transformation-based methods require (such as MCCA). Thus, this property is not only analytically elegant, but practically significant when \emph{c} is small compared with the feature space dimension (which includes emotion recognition, digit recognition, English character recognition, and many others), where \emph{c} ranges from a handful to a couple of dozens, but the feature space dimension could be hundreds or even thousands.\\\indent
In summary of the discussion so far, the information fusion algorithm based on DMCCA is given below:\\\indent
\textbf{Step 1}. Extract information from multimodal sources to form the training sample spaces.\\\indent
\textbf{Step 2}. Convert the extracted information into the normalized form and compute the matrixes \emph{C} and \emph{D}.\\\indent
\textbf{Step 3}. Compute the eigenvalues and eigenvectors of equation (19).\\\indent
\textbf{Step 4}. Obtain the fused information expression from equation (51), which is used for classification.\\

Next we will demonstrate that CCA, MCCA and DCCA are special cases of DMCCA.\\
1)	Relation with DCCA: when $P$=2, equation (19) turns into the following form:
\begin{equation} \ (C - D)\omega  = \rho D\omega \end{equation}
where
\begin{equation} \ C = \left[ \begin{array}{l}
{x_1}{x_1}^T\\
{x_2}A{x_1}^T
\end{array} \right.\left. \begin{array}{l}
{x_1}A{x_2}^T\\
{x_2}{x_2}^T
\end{array} \right] \end{equation}

\begin{equation} \ \;D = \left[ \begin{array}{l}
{x_1}{x_1}^T\\
0
\end{array} \right.\left. \begin{array}{l}
{\rm{  }}0\\
{x_2}{x_2}^T
\end{array} \right] \end{equation}
Thus, it transforms into the method of DCCA. Although DCCA also possesses this discriminative property, it can only deal with the mutual relationships between two random variables.\\

2)	Relation with MCCA: when \emph{A} is an identity matrix, the matrixes \emph{C} and \emph{D} of the DMCCA can be written:
\begin{equation} \ C = \left[ {\left( {\begin{array}{*{20}{c}}
{{x_1}{x_1}^T}& \ldots &{{x_1}{x_P}^T}\\
 \vdots & \ddots & \vdots \\
{{x_P}{x_1}^T}& \cdots &{{x_P}{x_P}^T}
\end{array}} \right)} \right] \end{equation}
\begin{equation} \ D = \left[ {\left( {\begin{array}{*{20}{c}}
{{x_1}{x_1}^T}& \ldots &0\\
 \vdots & \ddots & \vdots \\
0& \cdots &{{x_P}{x_P}^T}
\end{array}} \right)} \right] \end{equation}
It transforms into the method of MCCA.\\

3)	Relation with CCA: when $P$=2 and \emph{A} is an identity matrix, the matrixes \emph{C} and \emph{D} of the DMCCA can be described as:
\begin{equation} \ C = \left[ \begin{array}{l}
{x_1}{x_1}^T\\
{x_2}{x_1}^T
\end{array} \right.\left. \begin{array}{l}
{x_1}{x_2}^T\\
{x_2}{x_2}^T
\end{array} \right] \end{equation}
\begin{equation} \ D = \left[ \begin{array}{l}
{x_1}{x_1}^T\\
0
\end{array} \right.\left. \begin{array}{l}
{\rm{  }}0\\
{x_2}{x_2}^T
\end{array} \right] \end{equation}
It transforms into the method of CCA. Since there are no within-class and between-class correlation considered, CCA and MCCA do not possess the discriminative power as DMCCA. In addition, the best performance of CCA (and MCCA) is not predictable.
\section{Recognition By Multimodal/Multi-feature Fusion}
In this section, we examine the performance of DMCCA in several applications, ranging from multi-feature information fusion in handwritten digit recognition, audio-based and visual-based emotion recognition to multimodal information fusion in audiovisual-based human emotion recognition. A general block diagram of the proposed recognition system can be depicted as Fig.1.
\centerline {\includegraphics[height=3.8in,width=2.8in]{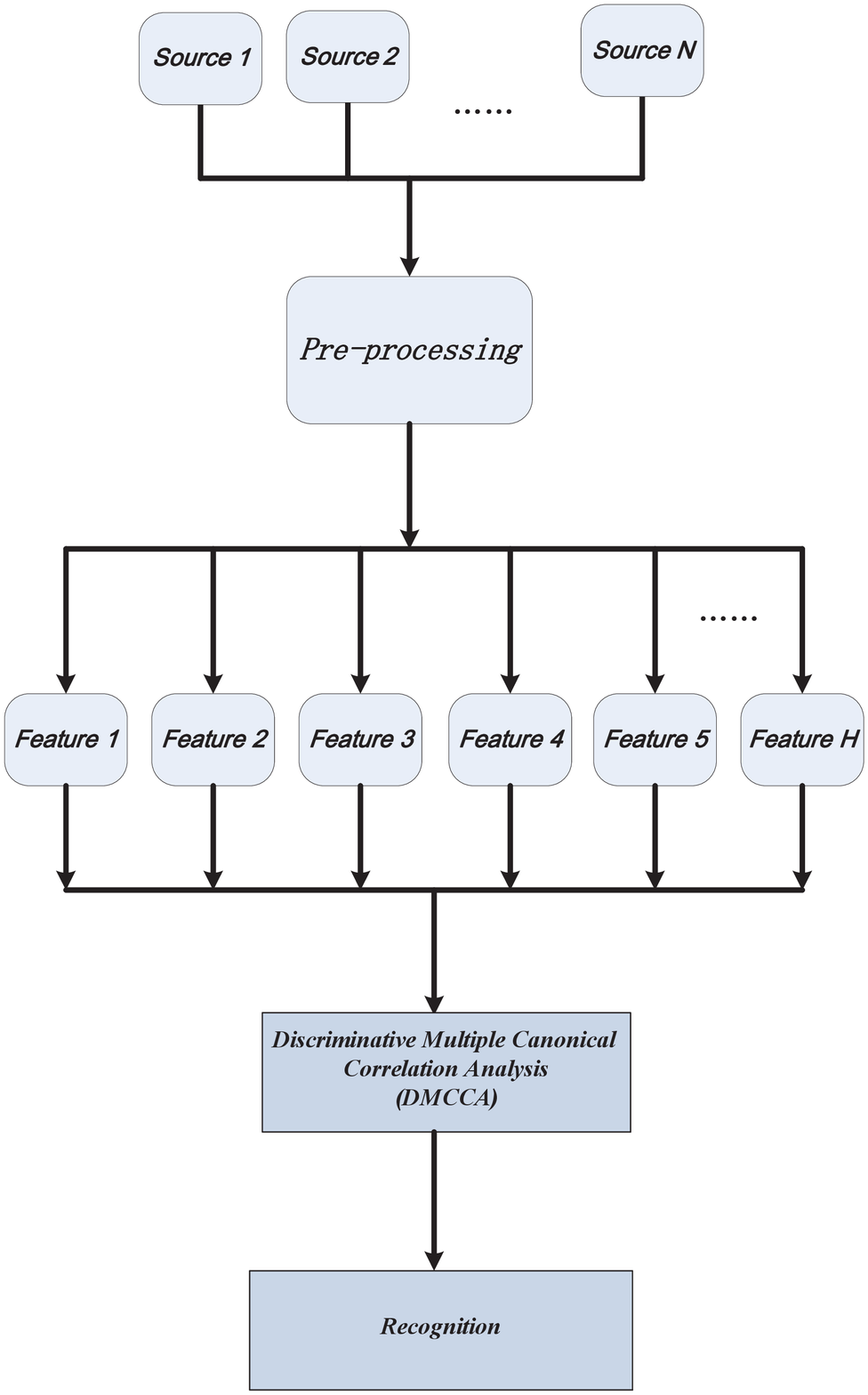}}\\ \ {Fig.1 The block diagram of DMCCA recognition system}\\
\subsection{The DMCCA for multi-feature fusion}
Multi-feature fusion is a special case of multimodal fusion. In multi-feature fusion, different sets of features are extracted from the same modality data but with different extraction methods, and thus highly likely carry complementary information. Therefore, the fusion of the multi-feature sets would lead to better recognition results.
\subsubsection{Handwritten Digit Recognition}
Handwritten digit recognition is an active research topic in OCR applications and pattern recognition/learning research. The most popular applications are postal mail sorting, bank check processing, form data entry, etc. While in information fusion communities, the problem of handwritten digit recognition has been used to test the performance of different algorithms. Since the performance of digit recognition largely depends on the quality of features, various feature extraction methods have been proposed, with Zernike moments [49] and Gabor-filter [50] being considered as two of the most popular and efficient. In this paper, three sets of features based on Zernike moments and Gabor-filter are extracted as follows:\\
1) 24-dimensional Gabor-filter feature with the mean of the digit images transformed by the Gabor filter [44].\\
2) 24-dimensional Gabor-filter feature with the standard deviation of the digit images transformed by the Gabor filter [44].\\
3) 36-dimensional Zernike moments feature [49].\\\
\centerline {\includegraphics[height=2.5in,width=1.3in]{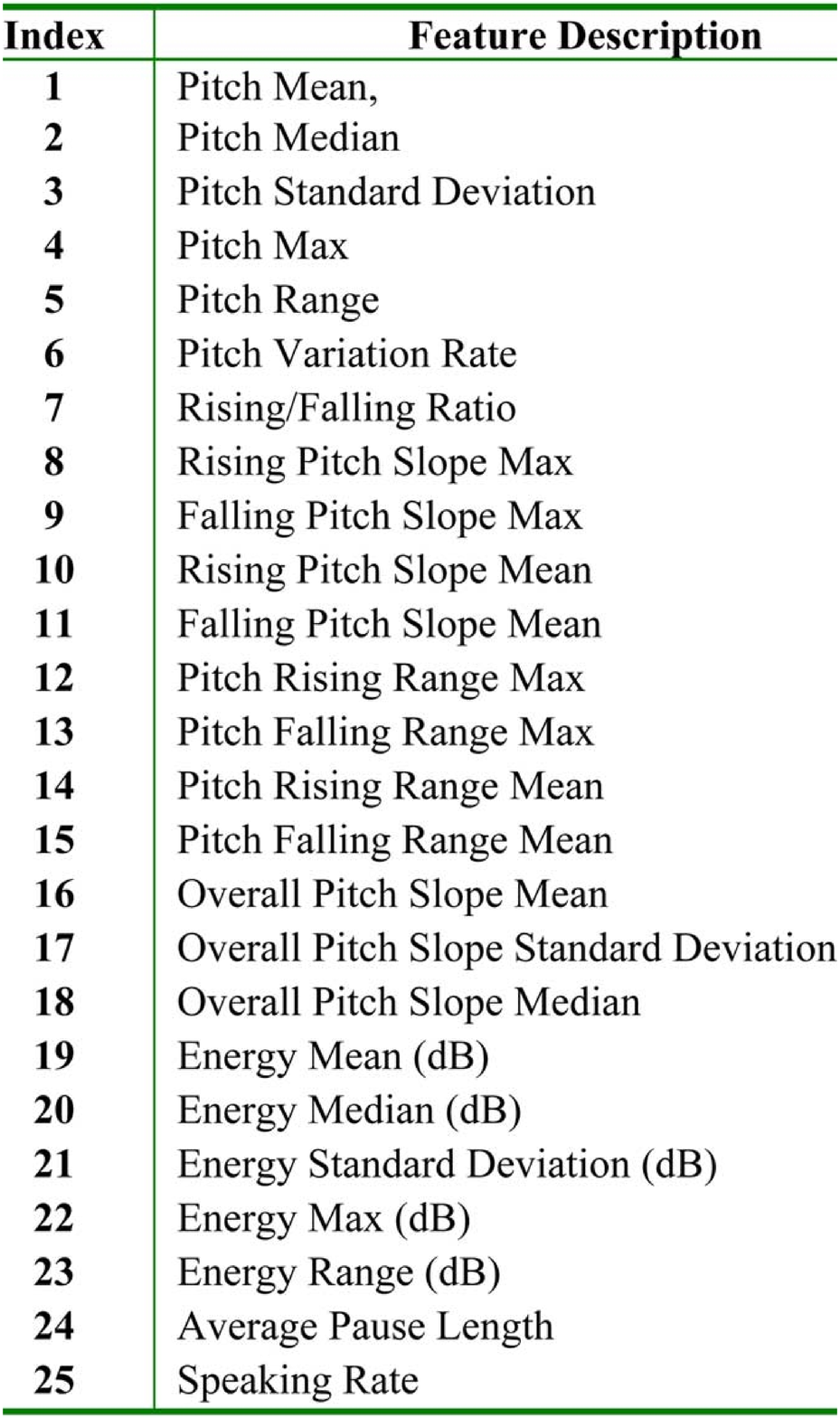}}\\ \centerline {Fig.2 Extracted prosodic features}\\\indent
\centerline {\includegraphics[height=2.3in,width=1.8in]{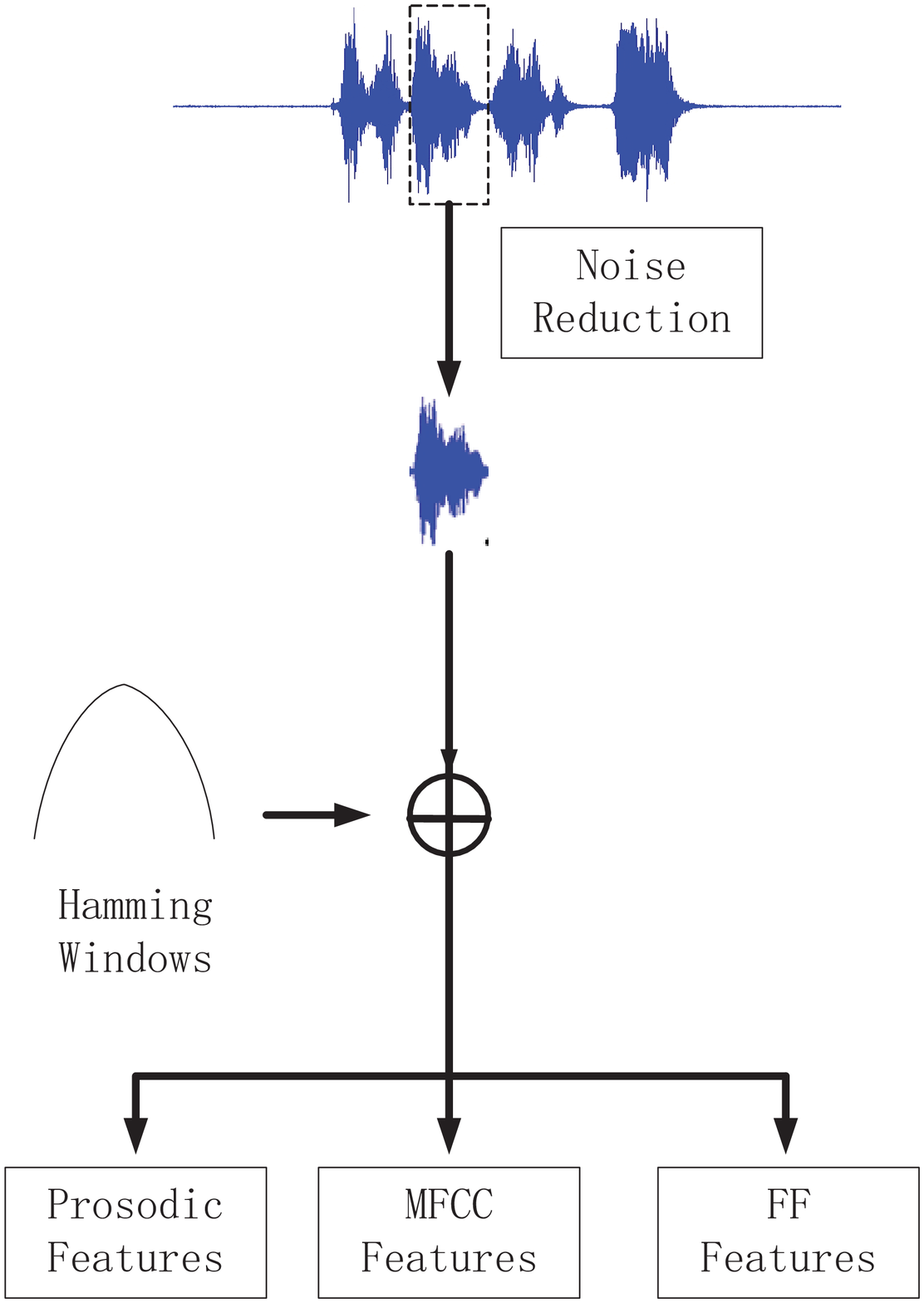}}\\ \centerline {Fig.3 Audio features extraction of emotion recognition}\\\indent
\subsubsection{Audio-based Emotion Recognition}
To build an emotion recognition system, the extraction of features that can truly represent the representative characteristics of the intended emotion is a challenge. For emotional audio, a good reference model is the human hearing system. Currently, Prosodic, MFCC and Formant Frequency (FF) are widely used in audio emotion recognition [8,12]. In this paper, we investigate the fusion of all these three types of features which are extracted as follows:\\\ 1)	Prosodic features as listed in Fig. 2.\\\ 2)	MFCC features: the mean, median, standard deviation, max, and range of the first 13 MFCC coefficients.\\\ 3)	Formant Frequency features: the mean, median, standard deviation, max and min of the first three formant frequencies.\\\indent The procedure of audio-based emotion recognition for feature extraction is shown in Fig.3. An input audio segment is first pre-processed by a wavelet coefficient thresholding method to reduce the recording machine and background noise [8]. To extract audio features, we perform spectral analysis which is only reliable when the signal is stationary. For audio, this holds only within the short time intervals of articulatory stability, during which a short time analysis can be performed by windowing a signal into a succession of frames. In this paper, we use a Hamming window of size 512 points, with 50 $ \% $ overlap between the adjacent frames.\\
\subsubsection{Visual-based Emotion Recognition}
Facial expression is another major factor in human emotion recognition. Existing solutions for facial expression analysis can be roughly categorized into two groups. One is to treat the human face as a whole unit [39-40], and the other is to represent the face by prominent components, such as mouth, eyes, nose, eyebrow, and chin [41-42]. Since focusing on only mouth, eyes, nose, eyebrow, and chin as facial components, the representation of the discriminative characteristics of human emotion might be inadequate. Therefore, in this work, we perform facial analysis by treating the face as a holistic pattern. The visual information is represented by Gabor wavelet features [43]. It allows description of spatial frequency structure in images while preserving information about spatial relations. In this paper, the Gabor filter bank is designed using the algorithm proposed in [44]. The designed Gabor filter bank consists of filters in 4 scales and 6 orientations. Apparently, the basic Gabor feature space is of high dimensionality, leading to high computational cost, and thus is not suitable for practical applications. We therefore take the mean, standard deviation and median of the magnitude of the transform coefficients of the filter banks as the three feature sets and fuse them for visual emotion recognition. The procedure of visual-based emotion recognition for features extraction is shown in Fig.4.
\centerline {\includegraphics[height=1.5in,width=2.0in]{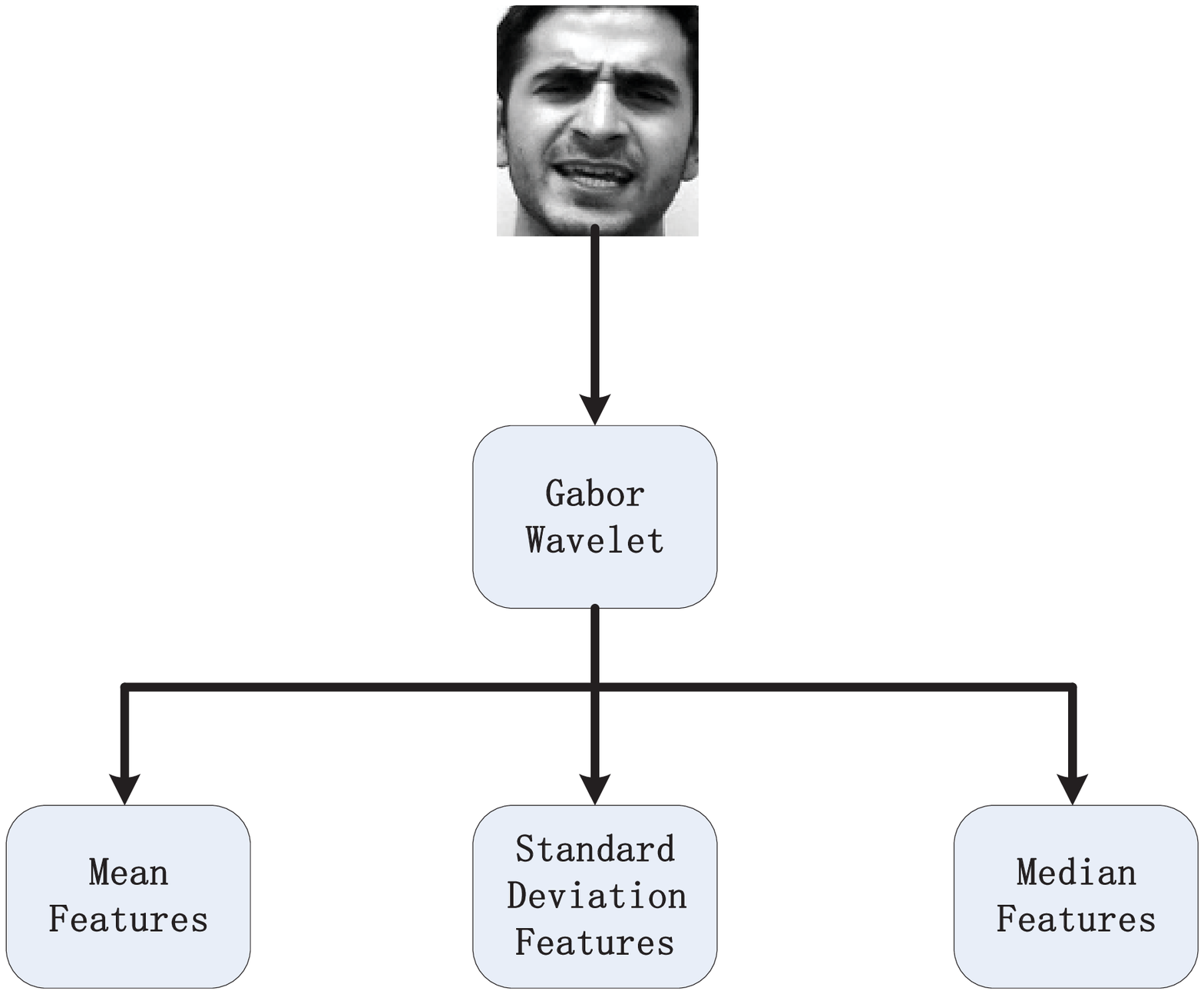}}\\  \centerline {Fig.4  Visual features extraction of emotion recognition}\\
\subsection{The DMCCA for multimodal fusion}
To evaluate the effectiveness of DMCCA in audiovisual emotion recognition, the Prosodic, MFCC, and Formant Frequency features of audio signals and the mean, standard deviation, median of the magnitude of the Gabor transform coefficients of images are combined as the audiovisual features for emotion recognition. Fig. 5, the realization of Fig. 1 for emotion recognition, depicts a block diagram of the proposed audiovisual-based emotion recognition system. The first part is the combination of Fig. 3 and Fig. 4, and the second part is DMCCA and recognition stages.
\begin{figure*}[t]
\centering
\includegraphics[height=2.0in,width=4.0in]{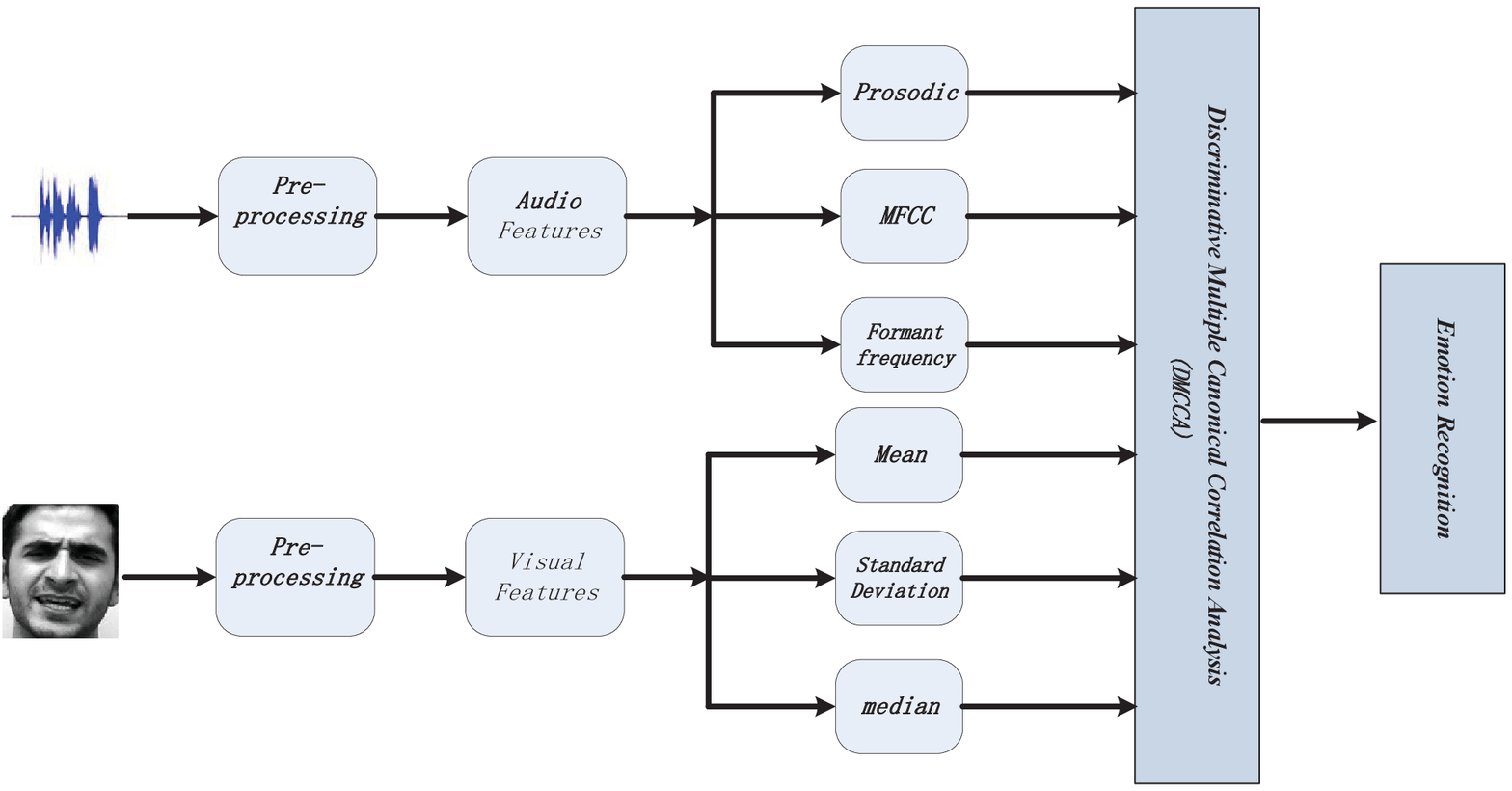}\\ Fig.5 The proposed multimodal information fusion for emotion recognition system \\
\end{figure*}

\subsection{Recognition algorithm analysis}
In terms of information fusion algorithm analysis, the method of FFS I in [23] is adopted to obtain the final
fused features for its better performance. Then we use the recognition algorithm proposed in [45], which can be written
as follows: \\\indent
Given two sets of features, represented by feature matrices
\begin{equation} \ {X^1} = [{x^1}_1,{x^1}_2,{x^1}_3,...{x^1}_d] \end{equation}
and
\begin{equation} \ {X^2} = [{x^2}_1,{x^2}_2,{x^2}_3,...{x^2}_d] \end{equation}
$dist[{X^1}{X^2}]$ is defined as
\begin{equation} \ dist[{X^1}{X^2}] = \sum\limits_{j = 1}^d {{{\left\| {{x^1}_j - {x^2}_j} \right\|}_2}} \end{equation}
where ${\left\| {a - b} \right\|_2}$ denotes the Euclidean distance between the two vectors \emph{a} and \emph{b}. \\\indent
Let the feature matrices of the \emph{N} training samples as ${F_1},{F_2},...{F_N}$ and each sample belongs to some class
${C_i}$ $ (i = 1,2...c )$, then for a given test sample \emph{I},
if
\begin{equation} \;dist[I,{F_l}] = \mathop {\min}\limits_j dist[I,{F_j}] (j = 1,2...N )\end{equation}
and
\begin{equation}{F_l} = {C_i}\end{equation}
the resulting decision is $I = {C_i}$. \\\indent
It is pointed out, in order to demonstrate the effectiveness of the proposed method, we also implemented the serial fusion, CCA, MCCA, DCCA for the purpose of comparison.
\section{Experimental Results and Analysis}
\subsection{The experimental results for multi-feature fusion}
\subsubsection{Handwritten Digit Recognition}
The MNIST database contains digits ranging from 0 to 9. Each digit is normalized and centered in a gray-level image with size 28$*$28. Some examples are shown in Fig. 6. In the experiments, we select 1500 samples to form the training subset and 1500 samples as the testing subsets, respectively. The performance of mean, standard deviation and zernike moment is first evaluated shown in TABLE I.\\
\centerline {\includegraphics[width=1.8in]{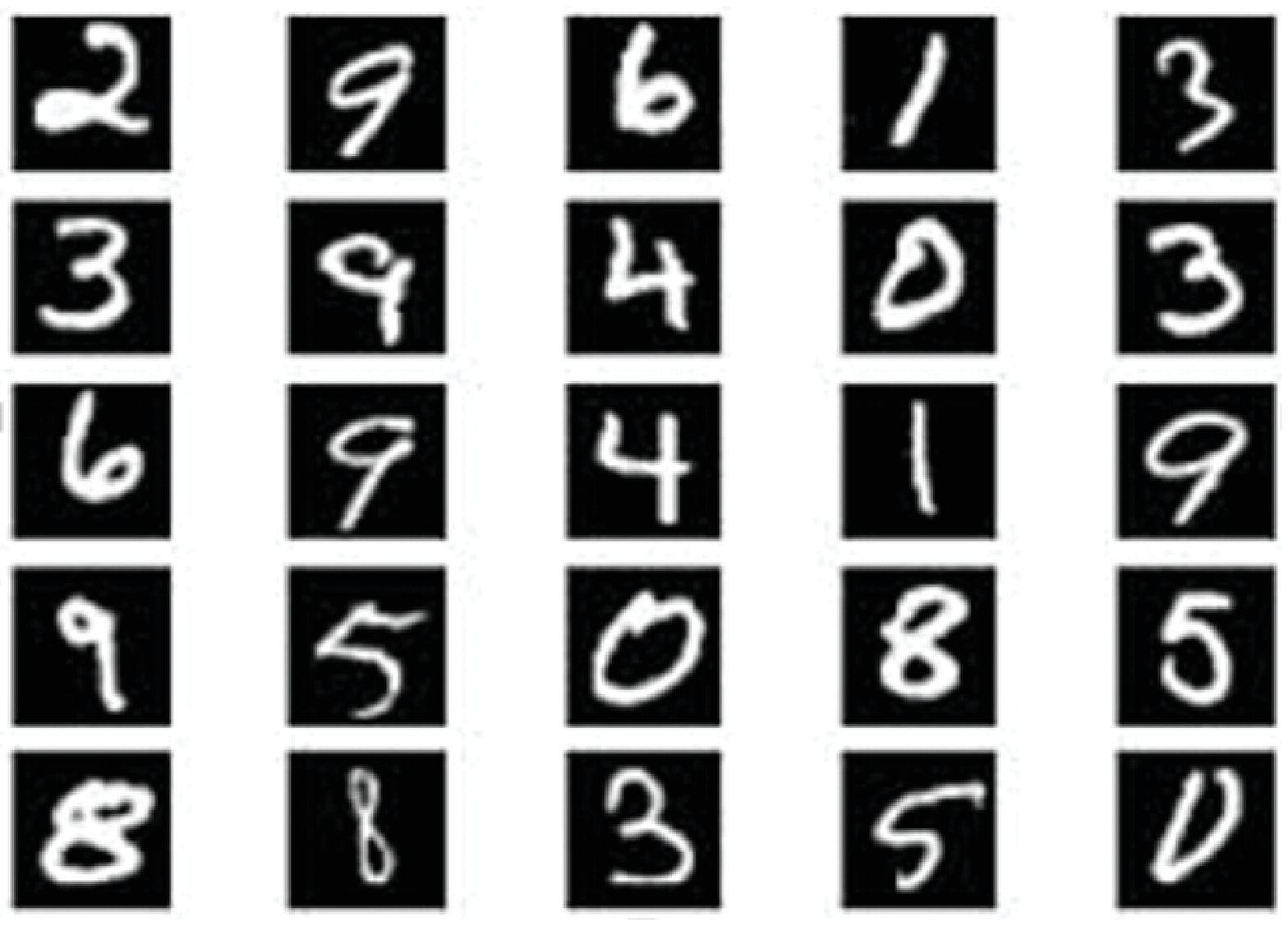}}\\  \centerline {Fig.6  Example images from the MNIST database}\\
\vspace*{-10pt}
\begin{table}[h]
\normalsize
\renewcommand{\arraystretch}{1.0}
\caption{\normalsize{Results of handwritten digit recognition with a single feature}}
\setlength{\abovecaptionskip}{0pt}
\setlength{\belowcaptionskip}{10pt}
\centering
\tabcolsep 0.073in
\begin{tabular}{cc}
\hline
Single Feature & Recognition Accuracy\\
\hline
Mean &49.13\%\\
Standard Deviation &52.60\%\\
Zernike &70.20\%\\
\hline
\end{tabular}
\end{table}\\

From TABLE I, the standard deviation (52.60\%) and zernike moment (70.20\%) features achieve better performance than the mean (49.13$\%$), and therefore will be used in CCA and DCCA which only take two sets of features. In addition, the performance based on the method of serial fusion with standard deviation \& zernike moment and all of the three features is implemented, respectively. The experimental results are shown in TABLE II.\\
\vspace*{-10pt}
\begin{table}[h]
\normalsize
\renewcommand{\arraystretch}{1.0}
\caption{\normalsize{Results of handwritten digit recognition by serial fusion}}
\setlength{\abovecaptionskip}{0pt}
\setlength{\belowcaptionskip}{10pt}
\centering
\tabcolsep 0.073in
\begin{tabular}{cc}
\hline
Serial Fusion & Recognition Accuracy\\
\hline
Standard Deviation \& Zernike  &70.20\%\\
All of the three features &70.33\%\\
\hline
\end{tabular}
\end{table}\\\indent
Next, the comparison among DMCCA, serial fusion, CCA, MCCA, and DCCA are implemented. 
The overall recognition rates are given in Fig. 7, with DMCCA providing the best performance, clearly showing the discriminative power of the DMCCA for information fusion in handwritten digit recognition. From the figure, it is clear that the application of DMCCA achieves the best performance when the projected dimension \emph{d} equals to 9 $<$ 10 = \emph{c}, the number of classes, confirming nicely with the mathematical proof in Section II.\\
\begin{figure*}[t]
\centering
\includegraphics[height=2.0in,width=5.7in]{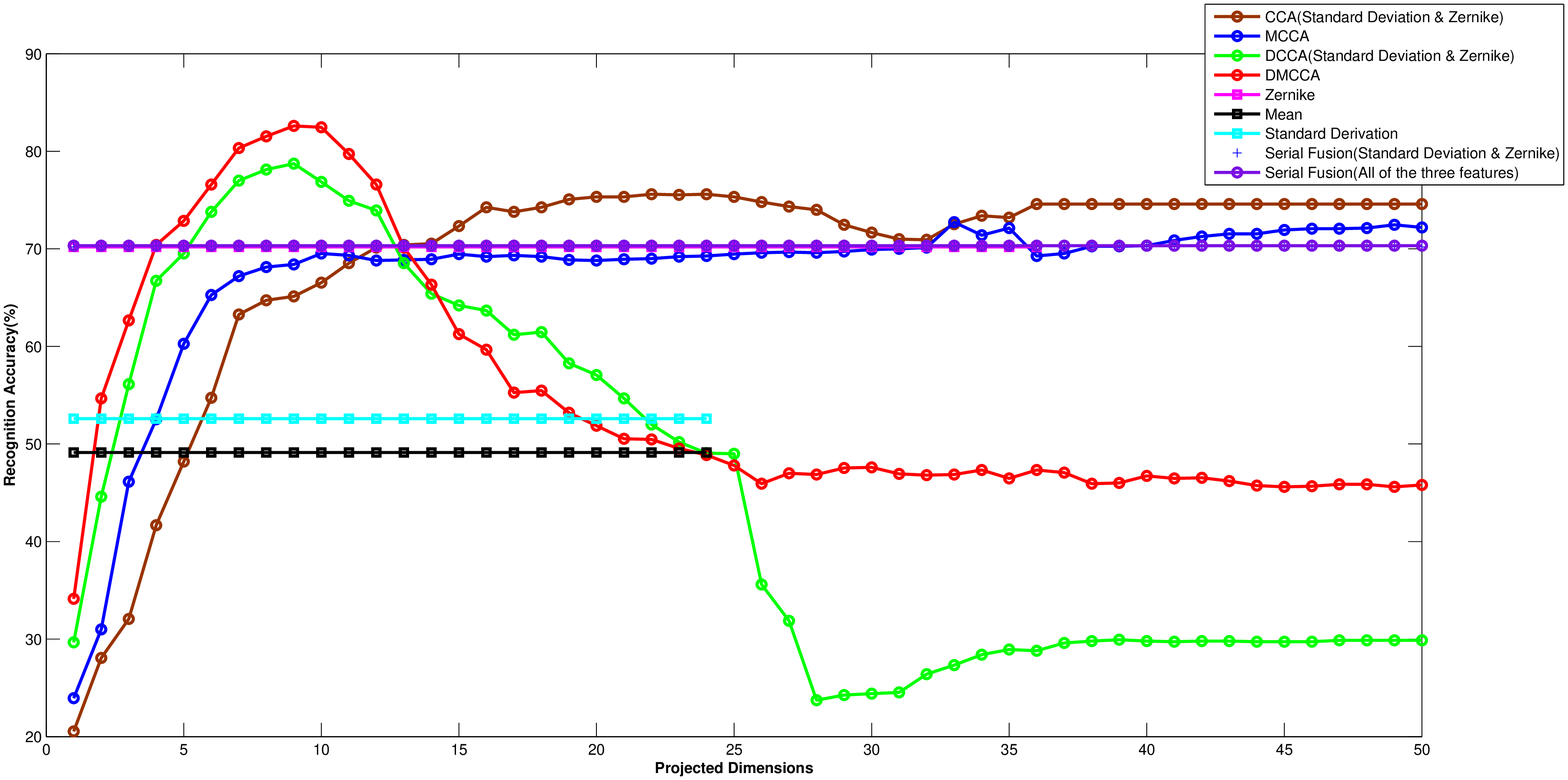}\\ Fig.7 Handwritten digit recognition experimental results of different methods on MNIST Database\\
\end{figure*}

\subsubsection{Audio-based Emotion Recognition}

To evaluate the performance of the proposed method, we conduct experiments on Ryerson Multimedia Lab (RML) and eNTERFACE (eNT) audiovisual databases[12], respectively. The RML database consists of video samples speaking six different languages (English, Mandarin, Urdu, Punjabi, Persian, and Italian) from eight subjects to express the six principal emotions-angry, disgust, fear, surprise, sadness and happiness. The frame rate for the video is 30 fps with audio recorded at a sampling rate of 22050 Hz. The size of the image frames is 720*480, and the face region has an average size of around 112*96.\\\indent
The eNTERFACE (eNT) database contains video samples from 43 subjects, also expressing the six basic emotions, with a sampling rate of 48000 Hz for audio channel and a video frame rate of 25 fps. The image frames have a size of 720*576, with the average size of the face region around 260*300. Example images from RML and eNTERFACE are shown in Fig. 8.
\begin{figure*}[t]
\centering
\includegraphics[height=1.8in,width=3.8in]{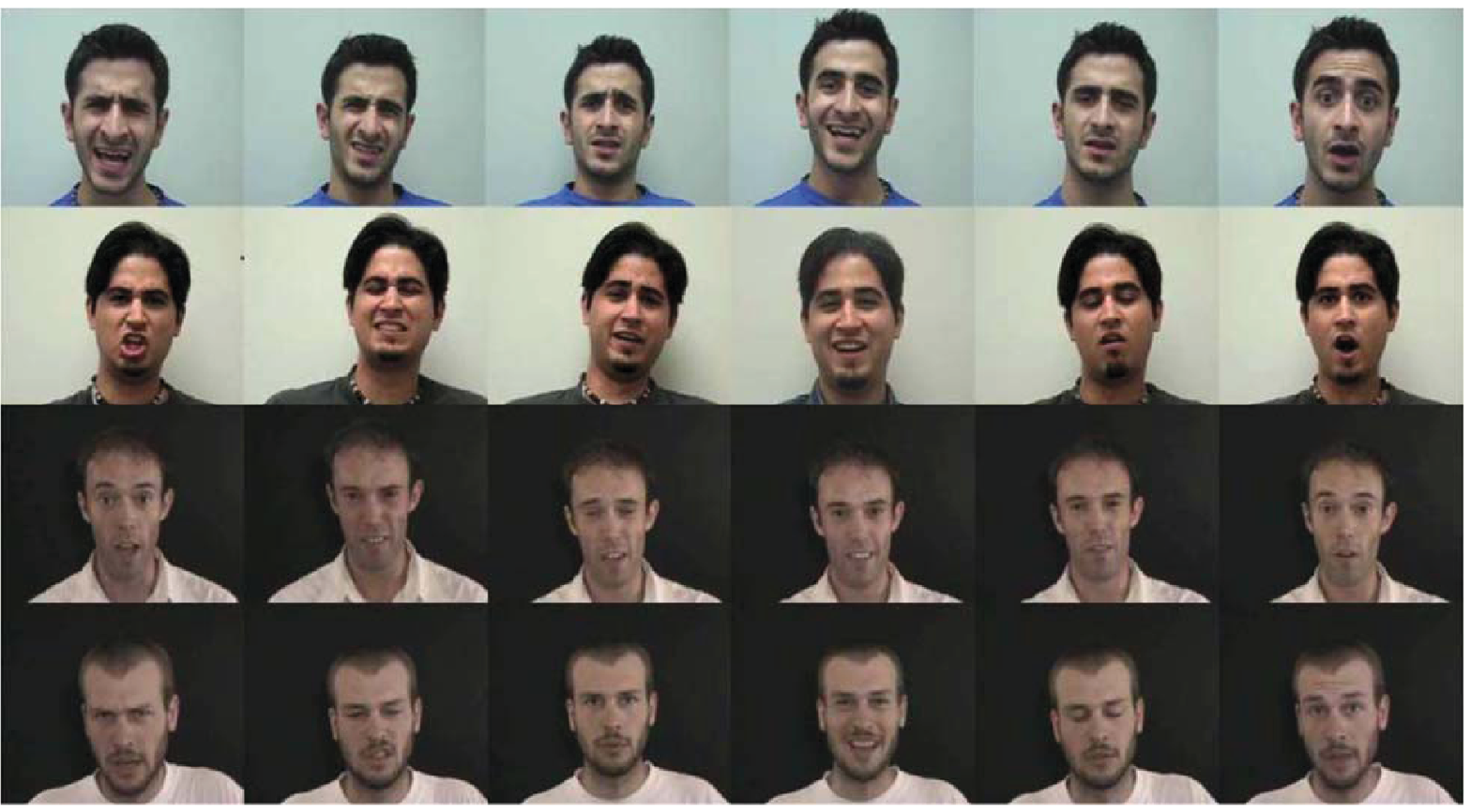}\\ Fig.8 Example facial expression images from the RML (Top two rows) and eNTERFACE (Bottom two rows) Databases\\
\end{figure*}

In the experiment, 456 audio samples of eight subjects from RML database and 456 audio samples of ten subjects from eNTERFACE database are selected, respectively. We divide both of the audio samples into training and testing subsets containing 360 and 96 samples each. As a benchmark, the performance of using Prosodic, MFCC and Formant Frequency features in emotion recognition are first evaluated, which are shown as TABLE \uppercase\expandafter{\romannumeral3}. The recognition accuracy is calculated as the ratio of the number of correctly classified samples over the total number of testing samples.\\
\vspace*{-10pt}
\begin{table}[h]
\normalsize
\renewcommand{\arraystretch}{1.0}
\caption{\normalsize{Results of Emotion Recognition With Single Audio Feature}}
\setlength{\abovecaptionskip}{0pt}
\setlength{\belowcaptionskip}{10pt}
\centering
\tabcolsep 0.073in
\begin{tabular}{cc}
\hline
Single Feature & Recognition Accuracy\\
\hline
Prosodic(RML) &53.13\%\\
MFCC(RML) &47.92\%\\
Formant Frequency(RML) &29.17\%\\
Prosodic(eNT) &55.21\%\\
MFCC(eNT) &39.58\%\\
Formant Frequency(eNT) &31.25\%\\
\hline
\end{tabular}
\end{table}\\

TABLE \uppercase\expandafter{\romannumeral3} suggests we should use the prosodic (53.13$\%$, 55.21$\%$) and MFCC (47.92$\%$, 39.58$\%$) features which perform the best individually, in CCA and DCCA which only need to take two sets of features. We also experimented on the method of serial fusion on RML and eNT databases with prosodic $\&$ MFCC features and all of the three features, respectively. The experimental results are shown in TABLE \uppercase\expandafter{\romannumeral4}.
\vspace*{-10pt}
\begin{table}[h]
\normalsize
\renewcommand{\arraystretch}{1.0}
\caption{\normalsize{The Experimental Results of Audio-Based Emotion Recognition with Serial Fusion}}
\setlength{\abovecaptionskip}{0pt}
\setlength{\belowcaptionskip}{10pt}
\centering
\tabcolsep 0.073in
\begin{tabular}{cc}
\hline
Serial Fusion & Recognition Accuracy\\
\hline
 Prosodic \& MFCC(RML)  &54.17\%\\
 All of the three features(RML) &44.79\%\\
 Prosodic \& MFCC(eNT)  &40.63\%\\
 All of the three features(eNT) &34.38\%\\
\hline
\end{tabular}
\end{table}\\

Then, we compare the performance of DMCCA with serial fusion, CCA, MCCA and DCCA. The overall recognition accurancies are shown in Fig. 9 and Fig. 10. Clearly, the discrimination power of the DMCCA provides a more effective modelling of the relationship between different features and achieves better performance than other methods.
\begin{figure*}[t]
\centering
\includegraphics[height=2.0in,width=5.6in]{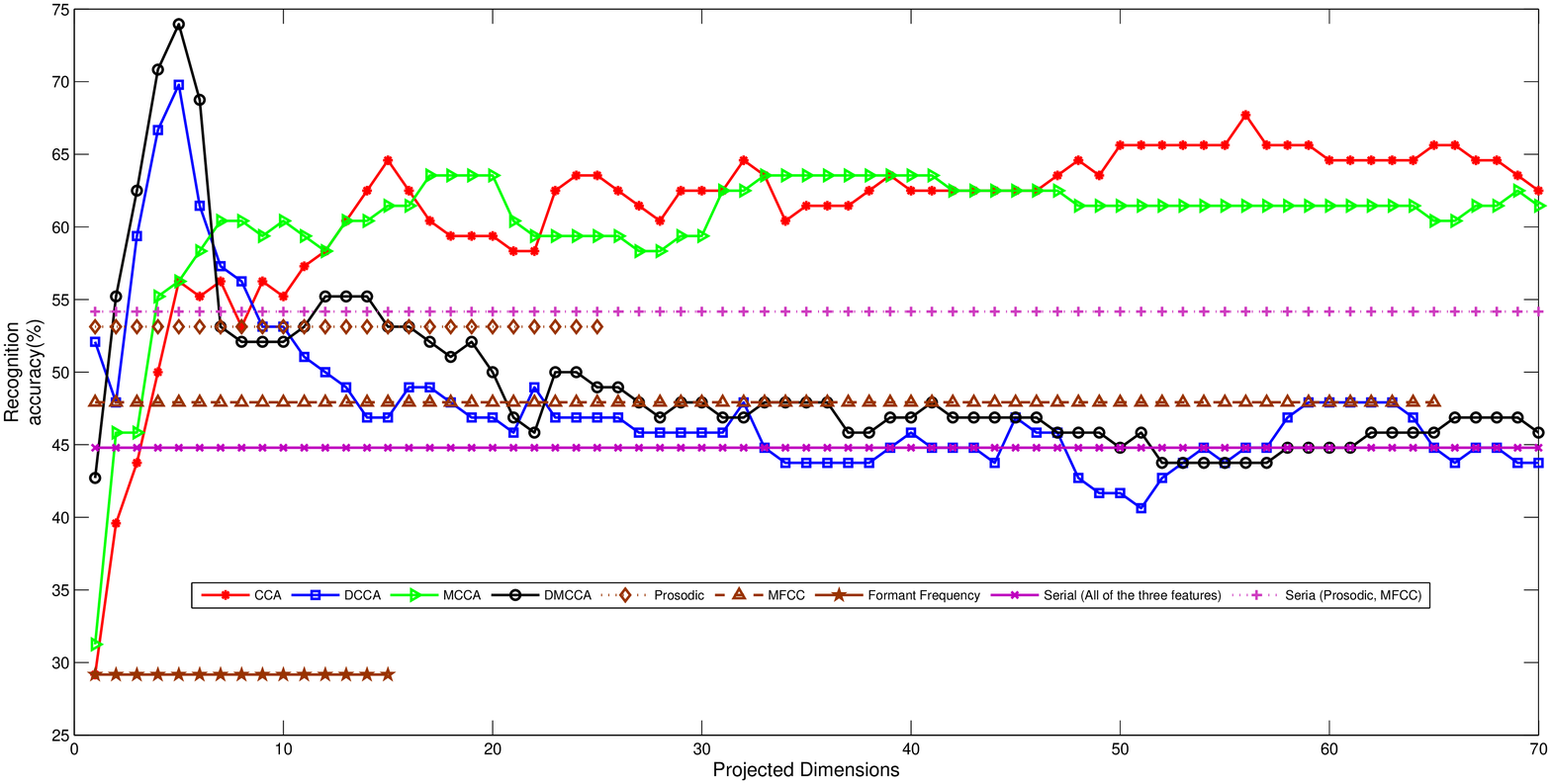}\\ Fig.9 Audio-based emotion recognition experimental results of different methods on RML Database\\
\end{figure*}
\begin{figure*}[t]
\centering
\includegraphics[height=2.0in,width=5.6in]{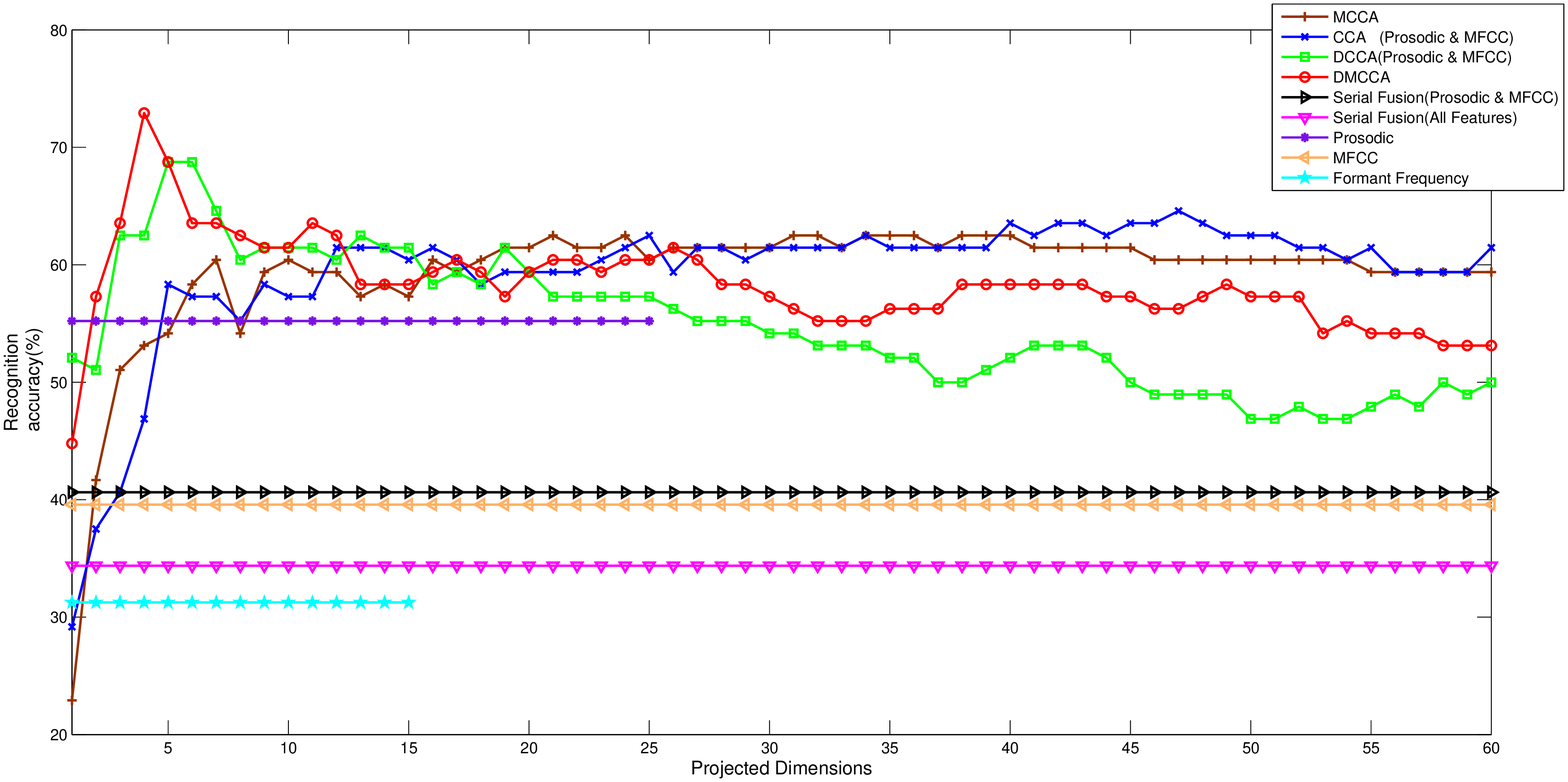}\\ Fig.10 Audio-based emotion recognition experimental results of different methods on eNTERFACE Database\\
\end{figure*}


\subsubsection{Visual-based Emotion Recognition}
To further evaluate the performance of DMCCA, we conduct experiments on RML and eNT visual emotion databases, respectively. Among the samples from RML database, 384 samples are chosen for training and 96 are chosen for evaluation. For eNT database, 456 visual samples are selected for experiment, training and testing subsets containing 360 and 96 each. As a benchmark, the performance of using mean, standard deviation and median features is evaluated. The results are shown as TABLE \uppercase\expandafter{\romannumeral5}.\\
\vspace*{-10pt}
\begin{table}[h]
\normalsize
\renewcommand{\arraystretch}{1.0}
\caption{\normalsize{Results of visual-based emotion recognition with single Gabor feature}}
\setlength{\abovecaptionskip}{0pt}
\setlength{\belowcaptionskip}{10pt}
\centering
\tabcolsep 0.073in
\begin{tabular}{cc}
\hline
Single Feature & Recognition Accuracy\\
\hline
Mean(RML) &60.42\%\\
Standard Deviation(RML) &67.71\%\\
Median(RML) &57.29\%\\
Mean(eNT) &75.00\%\\
Standard Deviation(eNT) &80.21\%\\
Median(eNT) &72.92\%\\
\hline
\end{tabular}
\end{table}\\

From TABLE \uppercase\expandafter{\romannumeral5}, it is observed that the features of mean (60.42$\%$, 75.00$\%$) and standard deviation (67.71$\%$, 80.21$\%$) achieve better performance in visual emotion recognition compared with the feature of median (57.29$\%$, 72.92$\%$). Thus, in the following experiments, we will use mean and standard deviation features in CCA and DCCA methods. The experiments using serial fusion with mean $\&$ standard deviation features and all of the three features are performed, which are shown in TABLE \uppercase\expandafter{\romannumeral6}.\\\indent

\vspace*{-10pt}
\begin{table}[h]
\normalsize
\renewcommand{\arraystretch}{1.0}
\caption{\normalsize{The Experimental Results of Visual-Based Emotion Recognition with Serial Fusion}}
\setlength{\abovecaptionskip}{0pt}
\setlength{\belowcaptionskip}{10pt}
\centering
\tabcolsep 0.07in
\begin{tabular}{cc}
\hline
Serial Fusion & Recognition Accuracy\\
\hline
Mean \& Standard Deviation(RML)  &69.79\%\\
 All of the three features(RML) &62.50\%\\
Mean \& Standard Deviation(eNT)  &79.17\%\\
 All of the three features(eNT) &80.21\%\\
\hline
\end{tabular}
\end{table}

The overall recognition results are illustrated in Fig. 11 and Fig.12, which shows, again, the proposed DMCCA outperforms the other methods.
\begin{figure*}[t]
\centering
\includegraphics[height=2.0in,width=5.6in]{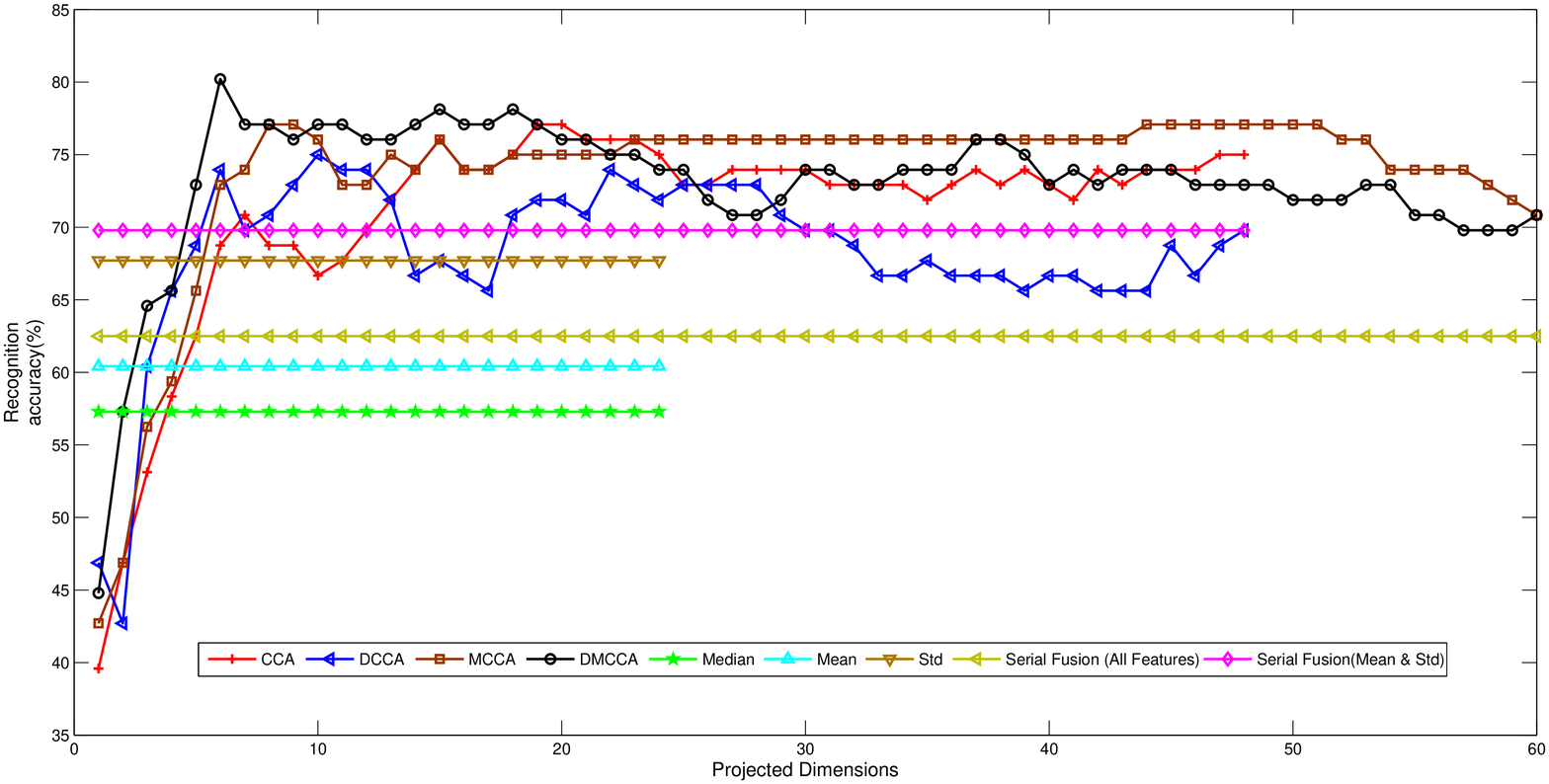}\\ Fig.11 Visual-based emotion recognition experimental results of different methods on RML Database\\
\end{figure*}

\begin{figure*}[t]
\centering
\includegraphics[height=2.0in,width=5.6in]{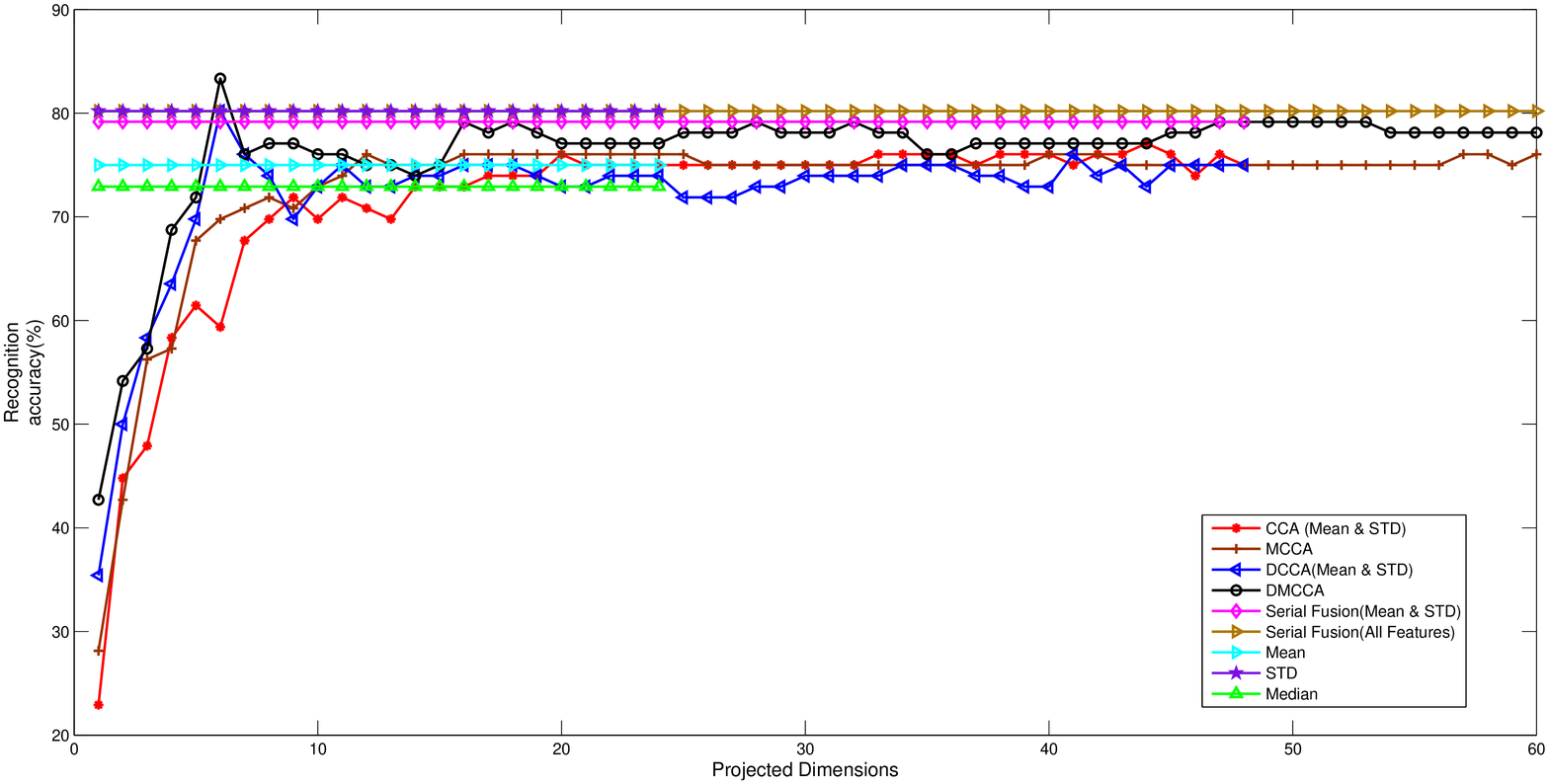}\\ Fig.12 Visual-based emotion recognition experimental results of different methods on eNTERFACE Database\\
\end{figure*}


\subsection{The experimental results for multimodal fusion}
For audiovisual-based emotion recognition, visual features are extracted from individual images. The planar envelope approximation method in the HSV color space is used for face detection [38]. To alleviate the high memory requirement of MCCA and DMCCA methods, as well as producing comparable results, 288 video clips of six basic emotions are selected for capturing the change of audio and visual information with respect to time simultaneously from RML database. Among them, 192 clips are chosen for training set and 96 for evaluation. For eNT database, 360 clips are chosen for training set and 96 for evaluation. From the previous experimental results, it is shown that the Prosodic features in audio and standard deviation of Gabor Transform coefficients in visual images could result in better performance in emotion recognition compared with other features. Therefore, in the following experiments, we will use Prosodic and standard deviation for the methods of CCA and DCCA in audiovisual-based multimodal fusion. Besides, the results of serial fusion on all the six audiovisual features are also investigated, and the overall recognition accuracy is 30.28$\%$ for RML database and 35.42$\%$ for eNT database. The performance by the methods of serial fusion, CCA, MCCA, DCCA, audio-based multi-feature DMCCA, visual-based multi-feature DMCCA and audiovisual-based DMCCA are shown in Fig. 13 and Fig. 14. From the experimental results, clearly, the discrimination power of the DMCCA provides a more effective modeling of the relationship between audiovisual information fusion.\\\indent In order to further demonstrate the effectiveness of DMCCA on multimodal fusion, we applied the method of DCCA with embedding (named as EDCCA) to bimodal emotion recognition and compared with DMCCA. There are six sets of features for bimodal emotion recognition. Three are audio: Prosodic, MFCC and Formant Frequency; and three are visual: Mean, Standard Deviation and Median calculated from Gabor wavelet transform. Naturally, we embed the three audio features together and the three visual features together, and perform EDCCA on the two embedded features. The overall recognition accuracies by EDCCA on RML and eNT datasets are shown in Fig. 15 and Fig. 16. \\\indent From Fig. 13 to Fig. 16, we observe the following in terms of optimal performance:\\
1) RML: 85\% (DMCCA, Green line in Fig. 13) $>$ 79\% (EDCCA) $>$ 69\% (DCCA, Blue line in Fig. 13)\\
2) eNT: 87\% (DMCCA, Green line in Fig. 14) $>$ 78\% (EDCCA) $>$ 77\% (DCCA, Blue line in Fig. 14)\\\indent
Therefore, properly embedding all features in two sets did improve the performance of DCCA in this example, but it is still inferior to the performance of DMCCA. Moreover, when the fusion involves features from three or more modalities, it is difficult, if not impossible, to design a reasonable embedding strategy. On the other hand, with a sound theoretical foundation, DMCCA can handle fusion involving any number of modalities.\\\indent From the above experimental results, it can be seen that the recognition accuracy of serial fusion is generally worse than CCA and related methods, and fusion does not help as shown in TABLE \uppercase\expandafter{\romannumeral2}, TABLE \uppercase\expandafter{\romannumeral4} and TABLE \uppercase\expandafter{\romannumeral6} justifying that simply putting the features from different channels together without considering the intrinsic structure and relationship results in low recognition accuracy.\\\indent On the other hand, the performance of DMCCA is much better than the methods compared in all cases. An important finding of the research is that, the exact location of optimal recognition performance occurs when the number of projected dimension \emph{d} is smaller than or equals to the number of classes, confirming nicely with the mathematical analysis presented in Section \uppercase\expandafter{\romannumeral2}. The significance here is that, we only need to calculate the first \emph{d} (is smaller than or equals to the number of classes) projected dimensions of DMCCA to obtain the desired recognition performance, eliminating the need of computing the complete transformation processes associated with most of the other methods, and thus substantially reducing the computational complexity to obtain the optimal recognition accuracy.
\section{CONCLUSIONS}
This paper proposed a new method for effective information fusion through the analysis of discriminative multiple canonical correlation (DMCCA). The key contribution of this work is that based on the proposed algorithm, we are able to verify that the best performance by discriminatory representation is achieved when the number of projected dimensions is smaller than or equals to the number of classes (\emph{c}) in the fused space. The significance of the contribution is that, only the first \emph{c} projected dimensions of the discriminative model need to be calculated to obtain the discriminatory representations, eliminating the need to compute the complete transformation process. This is a particularly attractive feature when dealing with large scale problems. Furthermore, we mathematically proved that CCA, MCCA and DCCA are special cases of the proposed DMCCA, unifying correlation based information fusion methods.\\\indent The proposed method is applied to multi-feature and multimodal information fusion problems in handwritten digit recognition and human emotion recognition. Experimental results demonstrate that the proposed method improves recognition performance with substantially reduced dimensionality of the feature space, leading to efficient practical pattern recognition.\\\indent There exist plenty of directions in which the current work can be extended. One of the worthwhile topics is to investigate kernelized version of the DMCCA based on the Kernel canonical correlation analysis (KCCA) [46] in order to obtain a more effective method to solve nonlinear problems in information fusion.

\begin{figure*}[t]
\centering
\includegraphics[height=2.0in,width=5.6in]{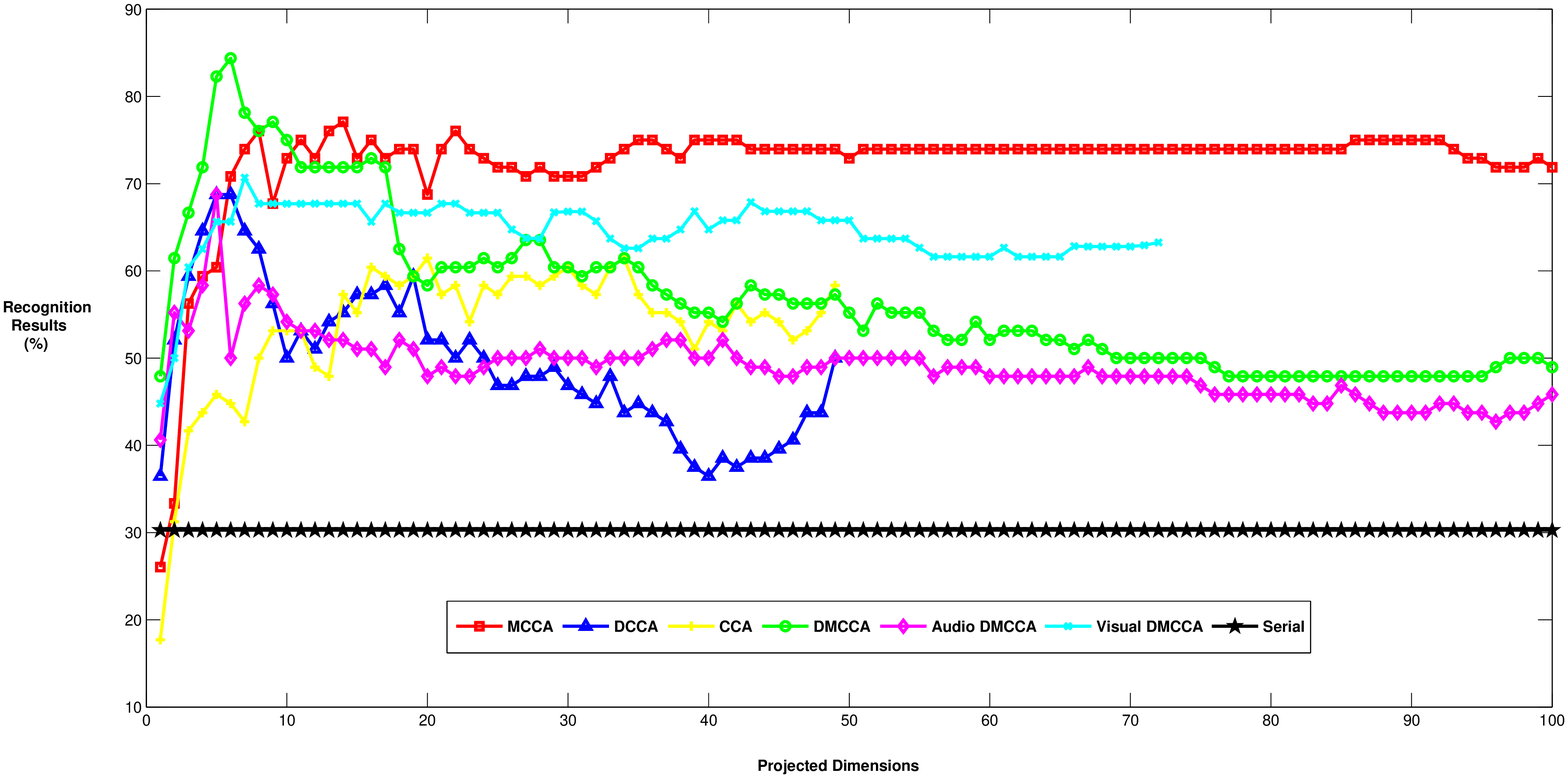}\\ Fig.13 Audiovisual emotion recognition experimental results by different methods on RML Database\\
\end{figure*}

\begin{figure*}[t]
\centering
\includegraphics[height=2.0in,width=5.6in]{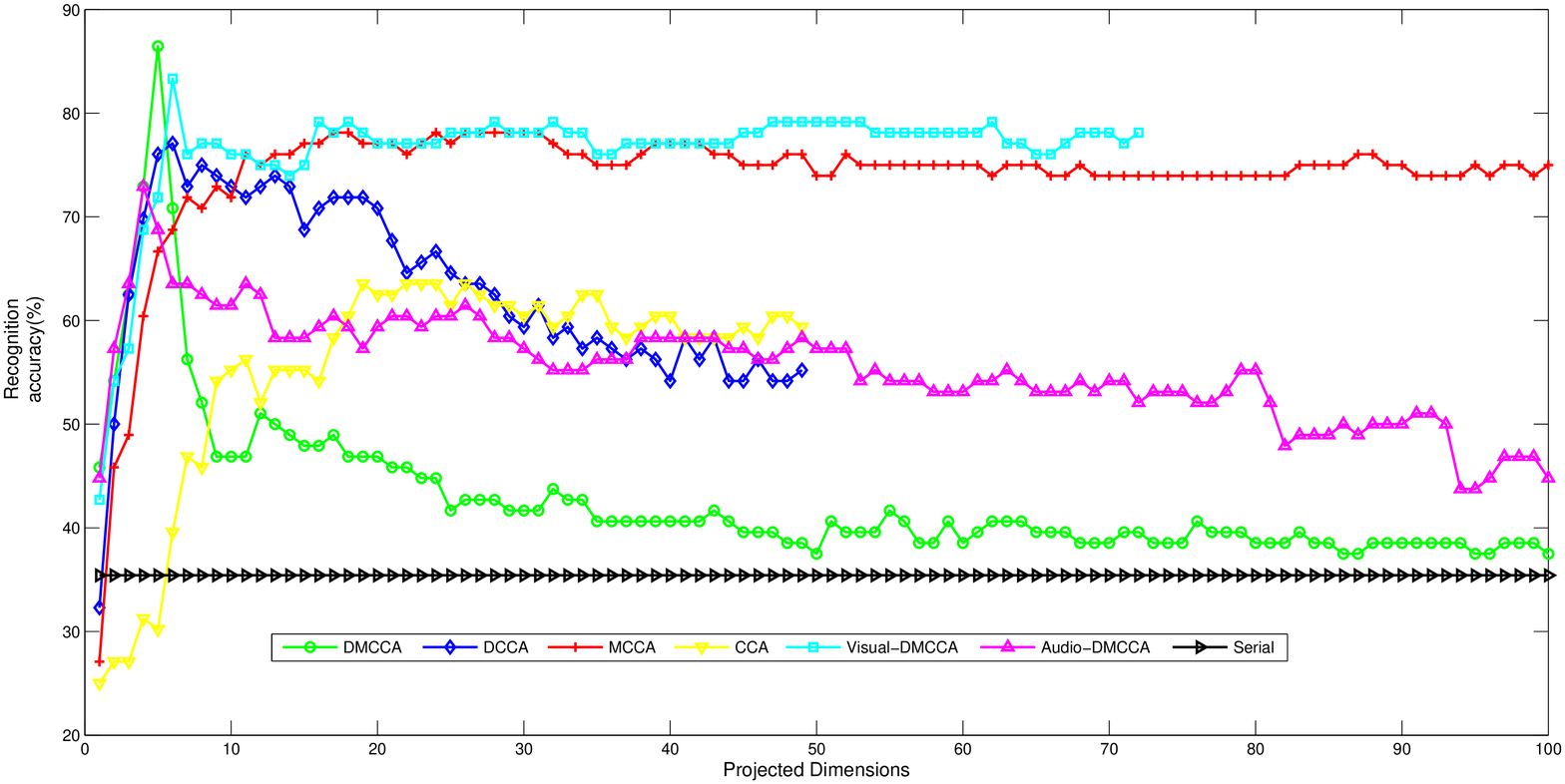}\\ Fig.14 Audiovisual emotion recognition experimental results by different methods on eNTERFACE Database\\
\end{figure*}

\begin{figure*}[t]
\centering
\includegraphics[height=2.0in,width=5.6in]{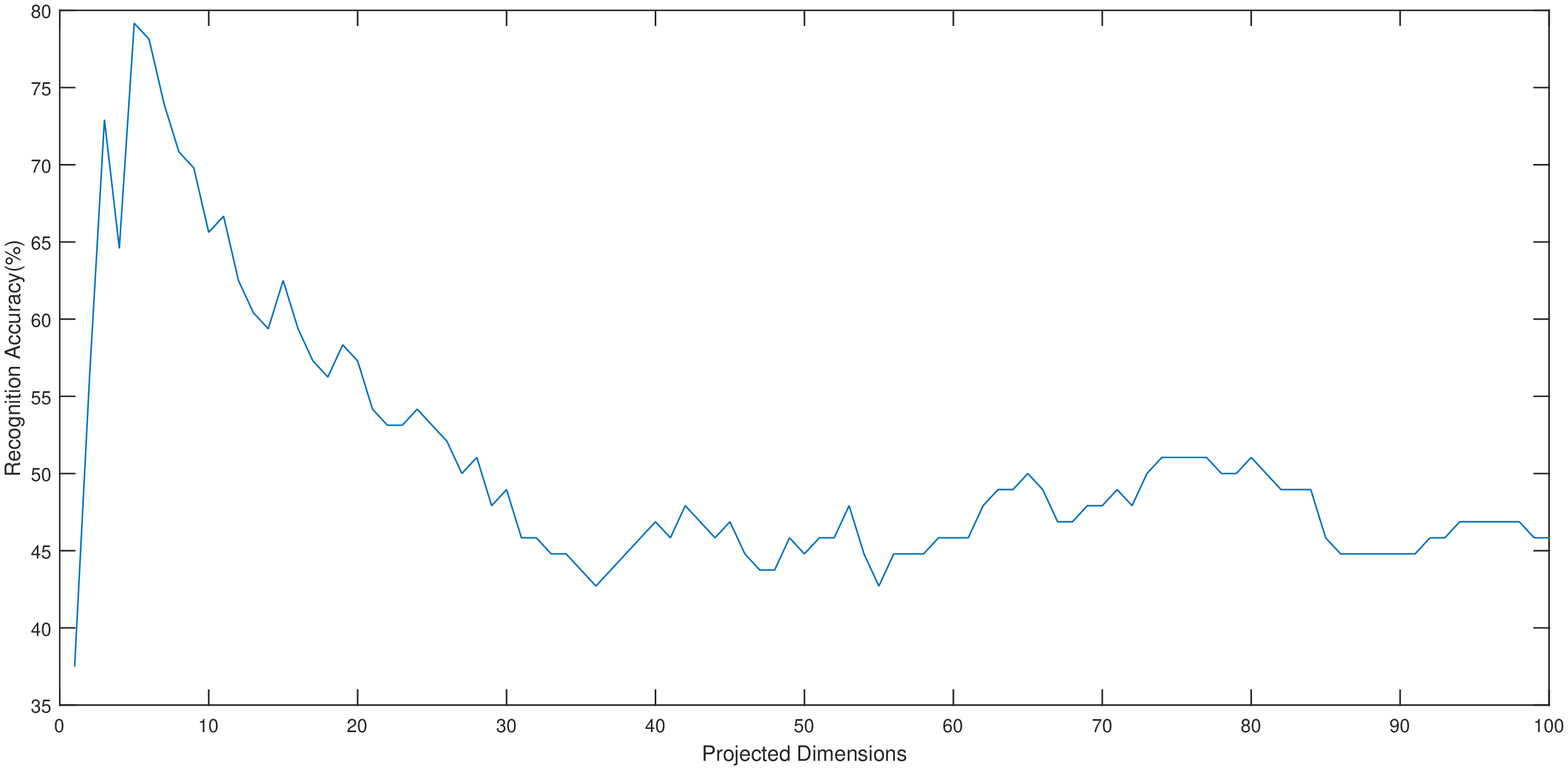}\\ Fig.15 Performance on audiovisual fusion on emotion recognition with the method of EDCCA
(RML Dataset)\\
\end{figure*}

\begin{figure*}[t]
\centering
\includegraphics[height=2.0in,width=5.6in]{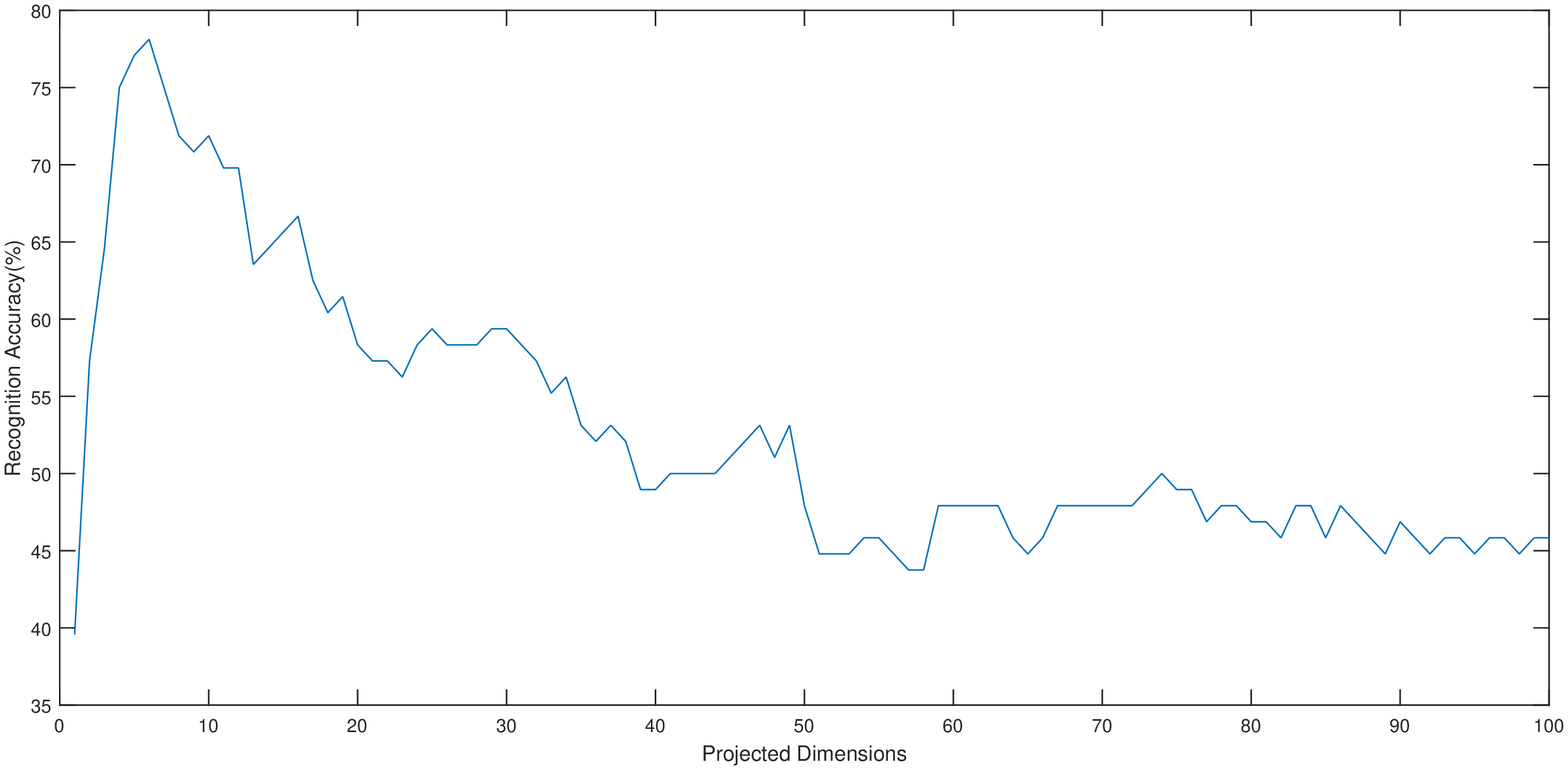}\\ Fig.16 Performance on audiovisual fusion on emotion recognition with the method of EDCCA
(eNT Dataset)\\
\end{figure*}

\section*{Appendix I}
Let ${x_i} = [{x_{i1}}^{(1)},{x_{i2}}^{(1)} \cdots {x_{i{n_1}}}^{(1)}, \cdots {x_{i1}}^{(c)},{x_{i2}}^{(c)} \cdots {x_{i{n_c}}}^{(c)}] \\ \in {R^{{m_i} \times n}}$, then
\begin{equation} \ e_{{n_{il}}} = {[\underbrace {0,0, \cdots 0,}_{\sum\limits_{u = 1}^{l - 1} {{n_{iu}}} }\underbrace {1,1, \cdots 1}_{{n_{il}}}\underbrace {0,0, \cdots 0}_{n - \sum\limits_{u = 1}^l {{n_{iu}}} }]^T} \in {{\bf{R}}^{\bf{n}}} \end{equation}

 \begin{equation} \ {\bf{1}} = {[1,1, \cdots 1]^{\bf{T}}} \in {{\bf{R}}^{\bf{n}}} \end{equation}
 where \emph{i} is the number sequence of the random features, \emph{n} is the total number of training samples, ${x_{ij}}^{(d)}$  denotes the \emph{j}th sample in the \emph{d}th class, respectively, and ${n_{il}}$ is the number of samples in the \emph{l}th class of ${x_i}$ set.

\begin{equation} \sum\limits_{l = 1}^c {{n_{il}}}  = n \end{equation}
where \emph{c} is the total number of classes. Note that, as the random features satisfies the property of zero-mean, it can be shown that:

\begin{equation} \ x_i  \bullet  {\bf{1}} = {\bf{0}} \end{equation}

Then, the within-class correlation matrix ${C_{{w_{{x_k}{x_m}}}}}$ can be written as:

\begin{equation} \begin{array}{l}
{C_{{w_{{x_k}{x_m}}}}} = \sum\limits_{l = 1}^c {\sum\limits_{h = 1}^{{n_{kl}}} {\sum\limits_{g = 1}^{{n_{ml}}} {{x_{kh}}^{(l)}{x_{mg}}^{(l)T}} } } \\
{\rm{    }} = \sum\limits_{l = 1}^c {({x_k}{e_{{n_{kl}}}}){{({x_m}{e_{{n_{ml}}}})}^T}} \\
{\rm{    }} = {x_k}A{x_m}^T
\end{array} \end{equation}

\begin{equation} \ A = \left[ {\left( {\begin{array}{*{20}{c}}{{H_{{n_{i1}} \times {n_{i1}}}}}& \ldots &0\\
 \vdots &{{H_{{n_{il}} \times {n_{il}}}}}& \vdots \\
0& \ldots &{{H_{_{{n_{ic}} \times {n_{ic}}}}}}
\end{array}} \right)} \right] \in {R^{n \times n}} \end{equation}
where ${H_{{n_{i{1}}} \times {n_{i1}}}}$ is in the form of ${n_{i1}} \times {n_{i1}}$ and all the elements in ${H_{{n_{i{1}}} \times {n_{i1}}}}$ are unit values. Similarly, the between-class correlation matrix ${C_{{b_{{x_k}{x_m}}}}}$ is in the form of:

\begin{equation} \begin{array}{l}
{C_{{b_{{x_k}{x_m}}}}} = \sum\limits_{l = 1}^c {\sum\limits_{\scriptstyle q = 1\hfill\atop
\scriptstyle l \ne q\hfill}^c {\sum\limits_{h = 1}^{{n_{kl}}} {\sum\limits_{g = 1}^{{n_{mq}}} {{x_{kh}}^{(l)}{x_{mg}}^{(q)T}} } } } \\
{\rm{    }} = \sum\limits_{l = 1}^c {\sum\limits_{q = 1}^c {\sum\limits_{h = 1}^{{n_{kl}}} {\sum\limits_{g = 1}^{{n_{mq}}} {{x_{kh}}^{(l)}{x_{mg}}^{(q)T} - } } } } \sum\limits_{l = 1}^c {\sum\limits_{h = 1}^{{n_{kl}}} {\sum\limits_{g = 1}^{{n_{ml}}} {{x_{kh}}^{(l)}{x_{mg}}^{(l)T}} } } \\
{\rm{    }} = ({x_k}{\bf{1}}){({x_m}{\bf{1}})^T} - {x_k}A{x_m}^T\\
{\rm{    }} =  - {x_k}A{x_m}^T
\end{array} \end{equation}

\section*{Acknowledgment}
This work is supported by the National Natural Science Foundation of China (NSFC, No.61071211), the State Key Program of NSFC (No. 61331021), the Key International Collaboration Program of NSFC (No. 61210005) and the Discovery Grant of Natural Science and Engineering Council of Canada (No. 238813/2010).


\begin{thebibliography}{1}


\bibitem{IEEEhowto:kopka}
L. Guan, Y. Wang, R. Zhang, Y. Tie, A. Bulzacki, and M. T. Ibrahim,``Multimodal information fusion for selected multimedia applications," \emph{Int. J. Multimedia Intell. Sec.}, vol. 1, no. 1, pp. 5-32, 2010.
\bibitem{IEEEhowto:kopka}
D. Roy, ``Grounding words in perception and action: Computational insights," \emph{Trends Cogn. Sci.}, vol. 9, no. 8, pp. 389-396, Aug. 2005.
\bibitem{IEEEhowto:kopka}
G. O. Deak, M. S. Bartlett, and T. Jebara, ``New trends in cognitive science: Integrative approaches to learning and development," \emph{Neurocomputing}, vol. 70, no. 13-15, pp. 2139-2147, Jan. 2007.
\bibitem{IEEEhowto:kopka}
A. K. Jain, A. Ross and S. Prabhakar, ``An Introduction to Biometric Recognition," \emph{IEEE Trans. on Circuits and Systems for Video Technology}, vol. 14, no. 1, pp. 4-20, Jan. 2004.
\bibitem{IEEEhowto:kopka}
K. C. Kwak, W. Pedrycz, ``Face recognition: A study in information fusion using fuzzy integral," \emph{Pattern Recognition Letters}, vol. 26, no. 6, pp. 719-733, 2005.
\bibitem{IEEEhowto:kopka}
S. Yildirim and S. Narayanan, ``Automatic Detection of Disfluency Boundaries in Spontaneous Speech of Children Using Audio-Visual Information," \emph{IEEE Trans on Audio, Speech, and Language Processing}, vol. 11, no. 1, pp. 2-12, Jan. 2009.
\bibitem{IEEEhowto:kopka}
H. Pan, S. E. Levinson, T. S. Huang and Z. Liang, ``A Fused Hidden Markov Model With Application to Bimodal Speech Processing,"  \emph{IEEE Trans. on  Signal Processing}, vol. 52, no. 3, pp. 573-581, Mar. 2004.
\bibitem{IEEEhowto:kopka}
Y. Wang and L. Guan, ``Recognizing human emotional state from audiovisual signals," \emph{IEEE Trans. Multimedia}, vol. 10, no. 5, pp. 936-946, Oct. 2008.
\bibitem{IEEEhowto:kopka}
Z. Zeng, M. Pantic, G. I. Roisman, and T. S. Huang, ``A survey of affect recognition methods: audio, visual, and spontaneous expressions," \emph{IEEE Trans. Pattern Anal. Mach. Intell.}, vol. 31, no. 1, pp. 39-58, 2008.
\bibitem{IEEEhowto:kopka}
T. Joshi, S. Dey, and D. Samanta, ``Multimodal biometrics: state of the art in fusion techniques," \emph{Int. J. Biomet.}, vol. 1, no. 4, pp. 393-417, July 2009.
\bibitem{IEEEhowto:kopka}
P. K. Atrey, M. A. Hossain, A. El Saddik, and M. S. Kankanhalli, ``Multimodal fusion for multimedia analysis: A survey," \emph{Multimedia Syst.}, vol. 16, no. 6, pp. 345-379, 2010.
\bibitem{IEEEhowto:kopka}
Y. Wang, L. Guan and A.N. Venetsanopoulos, ``Kernel based fusion with application to audiovisual emotion recognition," IEEE Trans. on Multimedia, vol. 14, no. 3, pp. 597-607, Jun 2012.
\bibitem{IEEEhowto:kopka}
D. Ruta and B. Gabrys. ``An overview of classifier fusion methods," \emph{Computing and Information systems}, vol. 7, no. 1, pp. 1-10, 2000.
\bibitem{IEEEhowto:kopka}
A. Ross and A. K. Jain, ``Information fusion in biometrics," \emph{Pattern Recognition Letter}, vol. 24, no. 13, pp. 2115-2125, 2003.
\bibitem{IEEEhowto:kopka}
Y.S. Huang, C.Y. Suen, ``Method of combining multiple experts for the recognition of unconstrained hand written numerals," \emph{IEEE Trans. Pattern Anal. Mach. Intell.}, vol. 7, no. 1, pp. 90-94, 1999.
\bibitem{IEEEhowto:kopka}
A.S. Constantinidis, M.C. Fairhurst, A.F.R. Rahman, ``A new multi-expert decision combination algorithm and its application to the detection of circumscribed masses in digital mammograms," \emph{Pattern Recognition}, vol. 34, no. 8, pp. 1528-1537, 2001.
\bibitem{IEEEhowto:kopka}
X.-Y. Jin, D. Zhang, J.-Y. Yang, ``Face recognition based on a group decision-making combination approach," \emph{Pattern Recognition}, vol. 36, no. 7, pp. 1675-1678, 2003.
\bibitem{IEEEhowto:kopka}
C.J. Liu, H. Wechsler, ``A shape-and texture-based enhanced Fisher classifier for face recognition," \emph{IEEE Trans. Image Processing}, vol. 10, no. 4, pp. 598-608, 2001.
\bibitem{IEEEhowto:kopka}
J. Yang, J.-Y. Yang, ``Generalized K-L transform based combined feature extraction," \emph{Pattern Recognition}, vol. 35, no. 1, pp. 295-297, 2002.
\bibitem{IEEEhowto:kopka}
J. Yang, J.Y. Yang, D. Zhang, J.F. Lu, ``Feature fusion: parallel strategy vs. serial strategy," \emph{Pattern Recognition}, vol. 36, no. 6, pp. 1369-1381, 2003.
\bibitem{IEEEhowto:kopka}
D.R. Hardoon, S. Szedmak and J.S. Taylor, ``Canonical correlation analysis: An overview with application to learning methods," \emph{Neural computation}, vol. 16, no. 12, pp.2639-2664, 2004.
\bibitem{IEEEhowto:kopka}
G. Lisanti, S. Karaman, I. Masi and A. Del Bimbo, ``Multi Channel-Kernel Canonical Correlation Analysis for Cross-View Person Re-Identification," \emph{IEEE Trans on Information Forensics and Security}, arXiv preprint arXiv:1607.02204, 2016.
\bibitem{IEEEhowto:kopka}
Q. Sun, S. Zeng, Y. Liu, P. Heng, and D. Xia, ``A new method of feature fusion and its application in image recognition," \emph{Pattern Recognition}, vol. 36, no. 12, pp. 2437-2448, Dec. 2005.
\bibitem{IEEEhowto:kopka}
H. Bredin and G. Chollet, ``Audio-visual speech synchrony measure for talking-face identity verification,"  \emph{IEEE International Conference on Acoustics, Speech and Signal Processing}, vol. 2, pp. 233-236, 2007.
\bibitem{IEEEhowto:kopka}
N. M. Correa, T. Adali, Y. Li, and V. D. Calhoun, ``Canonical correlation analysis for data fusion and group ingerences: examining applications of medical imaging data," \emph{IEEE Signal Process. Magazine}, vol. 27, no. 4, pp. 39-50, 2010.
\bibitem{IEEEhowto:kopka}
M. E. Sargin, Y. Yemez, E. Erzin, and A. M. Tekalp, ``Audiovisual synchronization and fusion using canonical correlation analysis," \emph{IEEE Trans. Multimedia}, vol. 9, no. 7, pp. 1396-1403, Nov. 2007.
\bibitem{IEEEhowto:kopka}
C. Fyfe and P. L. Lai, ``Kernel and nonlinear canonical correlation analysis," \emph{Int. J. Neural Syst.}, vol. 10, pp. 365-374, 2001.
\bibitem{IEEEhowto:kopka}
J. Zhao, Y. Fan, and W. Fan, ``Fusion of global and local features using KCCA for automatic target recognition," \emph{The Fifth International Conference on Image and Graphics}, pp. 958-962, 2009.
\bibitem{IEEEhowto:kopka}
X. Xu and Z. Mu, ``Feature fusion method based on KCCA for ear and profile face based multimodal recognition,"  \emph{2007 IEEE International Conference on Automation and Logistics}, pp. 620-623, 2007.
\bibitem{IEEEhowto:kopka}
M. B. Blaschko and C. H. Lampert, ``Correlation spectral clustering," \emph{ 2008 IEEE Conference on Computer Vision and Pattern Recognition}, pp. 1-8, 2008.
\bibitem{IEEEhowto:kopka}
M. J. Lyons, J. Budynek, et al., ``Classifying facial attributes using a 2-D Gabor wavelet representation and discriminant analysis," \emph{The Fourth IEEE International Conference on Automatic Face and Gesture Recognition}, pp. 202-207, March 2000.
\bibitem{IEEEhowto:kopka}
T.K. Sun, S.C. Chen, Z. Jin, J.Y. Yang, ``Kernelized Discriminative Canonical Correlation Analysis," \emph{International Conference on Wavelet Analysis and Pattern Recognition}, vol. 2, pp. 1283-1287, 2007.
\bibitem{IEEEhowto:kopka}
T. K. Kim, J. Kittler and R. Cipolla, ``Discriminative learning and recognition of image set classes using canonical correlations," \emph{IEEE Trans on Pattern Analysis and Machine Intelligence}, Vol. 29, No. 6, pp.1005-1018, 2007.
\bibitem{IEEEhowto:kopka}
O. Arandjelovic, ``Discriminative extended canonical correlation analysis for pattern set matching," \emph{Machine Learning}, pp. 1-18, 2013.
\bibitem{IEEEhowto:kopka}
A. A. Nielsen, ``Multiset Canonical Correlations Analysis and Multispectral, Truly Multitemporal Remote Sensing Data," \emph{IEEE Trans. Image Processing}, vol. 11, no. 3, pp. 293-305, Mar. 2002.
\bibitem{IEEEhowto:kopka}
T. Adali, W. Wang, V.D. Calhoun, ``Joint Blind Source Separation by Multiset Canonical Correlation Analysis," \emph{IEEE Trans. on  Signal Processing}, vol. 57, no. 10, pp. 3918-3929, Oct. 2009.
\bibitem{IEEEhowto:kopka}
J. Via, I. Santamara, d J. Pérez, ``Canonical correlation analysis (CCA) algorithms for multiple data sets: application to blind SIMO equalization," \emph{13th European Signal Processing Conference}, pp. 1-4, 2005.
\bibitem{IEEEhowto:kopka}
K. Praveen, S. Makrogiannis, and N. Bourbakis, ``A survey of skin-color modeling and detection methods," \emph{Pattern recognition}, vol. 40, no.3, pp. 1106-1122, 2007.
\bibitem{IEEEhowto:kopka}
M. J. Lyons, J. Budynek, A. Plante, and S. Akamatsu, ``Classifying facial attributes using a 2-D Gabor wavelet representation and discriminant analysis," \emph{The 4th International Conference on Automatic Face and Gesture
Recognition}, pp. 202-207, 2000.
\bibitem{IEEEhowto:kopka}
L. Ma and K. Khorasani, ``Facial expression recognition using constructive feedforward neural networks," \emph{IEEE Trans. Syst., Man, Cybern B: Cybern}, vol. 34, pp. 1588-1595, 2004.
\bibitem{IEEEhowto:kopka}
M. Pantic and L. J. M. rothkrantz, ``Facial action recognition for facial expression analysis from static face image", \emph{IEEE Trans. Syst., Man, Cybern B:Cybern}, vol. 34, pp. 1449-1461, 2004.
\bibitem{IEEEhowto:kopka}
L. C. De Silva and S. C. Hui, ``Real-time facial feature extraction and emotion recognition," \emph{The 4th International Conference on Information, Communications and Signal Processing}, vol. 3, pp. 1310-1314, 2003.
\bibitem{IEEEhowto:kopka}
M. J. Lyons, J. Budynek, A. Plante, and S. Akamatsu, ``Classifying facial attributes using a 2-D Gabor wavelet representation and discriminant analysis," \emph{The 4th Int. Conf. Automatic Face and Gesture Recognition}, pp. 202-207, 2000.
\bibitem{IEEEhowto:kopka}
B. S. Manjunath and W. Y. Ma, ``Texture features for browsing and retrieval of image data", \emph{IEEE Trans. Pattern Anal. Machine Intell.}, vol. 18, pp. 837-842, Aug. 1996.
\bibitem{IEEEhowto:kopka}
B. H. Shekar, M. S. Kumari, L. M. Mestetskiy and N. F. Dyshkant,``Face recognition using kernel entropy component analysis," \emph{Neurocomputing}, vol. 74, no. 6, pp. 1053-1057, 2011.
\bibitem{IEEEhowto:kopka}
W.M. Zheng, X.Y Zhou, C.R Zou, L. Zhao, ``facial expression recognition using kernel canonical correlation analysis (KCCA)," \emph{IEEE Trans. on Neural Networks}, Vol. 17, No. 1, pp. 233-238, 2006.
\bibitem{IEEEhowto:kopka}
J.L. Hu, J.W. Wen, J.S Yuan, Y.P. Tan, ``Large margin multi-metric learning for face and kinship verification in the wild", \emph{The 12th Asian Conference on Computer Vision}, pp. 1-16, 2014.
\bibitem{IEEEhowto:kopka}
G. Andrew, R. Arora, J. Bilmes, K. Livescu, ``Deep Canonical Correlation Analysis," \emph{The Proceedings of the 30th International Conference on Machine Learning}, pp. 1247-1255. 2013.
\bibitem{IEEEhowto:kopka}
K. Alireza, H.Yawhua, ``Invariant image recognition by Zernike moments," \emph{IEEE Trans. Pattern Anal. Mach. Intell.}, no. 12, pp. 489-497, 1990.
\bibitem{IEEEhowto:kopka}
Y. Hamamoto, S. Uchimura, M. Watanabe, T. Yasuda and S. Tomita, ``Recognition of handwritten numerals using Gabor features," \emph{International Conference on Pattern Recognition}, vol. 3, pp. 250-250, 1996.
\bibitem{IEEEhowto:kopka}
S. Zheng, X. Cai, C.H. Ding, F. Nie and H. Huang, ``A Closed Form Solution to Multi-View Low-Rank Regression," \emph{2015 The Twenty-Ninth AAAI Conference on Artificial Intelligence}, pp. 1973-1979, 2015.
\bibitem{IEEEhowto:kopka}
C. Xiao, F. Nie, W. Cai and H. Huang, ``Heterogeneous image features integration via multi-modal semi-supervised learning model," \emph{IEEE International Conference on Computer Vision}, pp. 1737-1744. 2013.
\bibitem{IEEEhowto:kopka}
W. Hua, F. Nie, H. Huang and C. Ding, ``Heterogeneous visual features fusion via sparse multimodal machine," \emph{IEEE Conference on Computer Vision and Pattern Recognition}, pp. 3097-3102, 2013.
\bibitem{IEEEhowto:kopka}
C. Hou, C. Zhang, Y. Wu and F. Nie, ``Multiple view semi-supervised dimensionality reduction," \emph{Pattern Recognition}, vol. 43, no. 3, pp. 720-730, 2010.
\bibitem{IEEEhowto:kopka}
 A. Sharma, A. Kumar, H. Daume and D.W. Jacobs, ``Generalized multiview analysis: A discriminative latent space,"  \emph{2012 IEEE Conference on Computer Vision and Pattern Recognition (CVPR)}, pp. 2160-2167, 2012.
\bibitem{IEEEhowto:kopka}
M. Kan, S. Shan, H. Zhang, S. Lao and X. Chen, ``Multi-view discriminant analysis,"  \emph{2012 European Conference on Computer Vision}, pp. 808-821, 2012.
\bibitem{IEEEhowto:kopka}
M.E. Sargent, Y. Yemez, E. Erzin and A.M. Tekalp, ``Audiovisual synchronization and fusion using canonical correlation analysis," \emph{IEEE Transactions on Multimedia}, vol.9, no. 7, pp.1396-1403, 2007.
\end{thebibliography}
\end{document}